\documentclass{article}

\usepackage{graphicx}
\usepackage{booktabs} 
\usepackage{nicefrac}       
\usepackage{microtype}
\usepackage{subcaption}
\usepackage{hyperref}
\usepackage{amsmath}
\usepackage{amsfonts}
\usepackage{enumerate}
\usepackage{bm}

\newcommand{\beginsupplement}{%
        \setcounter{table}{0}
        \renewcommand{\thetable}{S\arabic{table}}%
        \setcounter{figure}{0}
        \renewcommand{\thefigure}{S\arabic{figure}}%
     }

\usepackage[accepted]{icml2019}

\icmltitlerunning{Policy Consolidation for Continual Reinforcement Learning}

\begin{document}

\twocolumn[
\icmltitle{Policy Consolidation for Continual Reinforcement Learning}




\begin{icmlauthorlist}
\icmlauthor{Christos Kaplanis}{impdoc,impbio}
\icmlauthor{Murray Shanahan}{impdoc,deep}
\icmlauthor{Claudia Clopath}{impbio}
\end{icmlauthorlist}

\icmlaffiliation{impdoc}{Department of Computing, Imperial College London}
\icmlaffiliation{deep}{DeepMind, London}
\icmlaffiliation{impbio}{Department of Bioengineering, Imperial College London}

\icmlcorrespondingauthor{Christos Kaplanis}{christos.kaplanis14@imperial.ac.uk}

\icmlkeywords{Machine Learning, ICML}

\vskip 0.3in
]



\printAffiliationsAndNotice{} 

\begin{abstract}
We propose a method for tackling catastrophic forgetting in deep reinforcement learning that is \textit{agnostic} to the timescale of changes in the distribution of experiences, does not require knowledge of task boundaries, and can adapt in \textit{continuously} changing environments. In our \textit{policy consolidation} model, the policy network interacts with a cascade of hidden networks that simultaneously remember the agent's policy at a range of timescales and regularise the current policy by its own history, thereby improving its ability to learn without forgetting. We find that the model improves continual learning relative to baselines on a number of continuous control tasks in single-task, alternating two-task, and multi-agent competitive self-play settings.
\end{abstract}

\section{Introduction}
\label{submission}

An agent that continuously builds on its experiences to develop increasingly complex skills should be able to adapt quickly while also preserving its previously acquired knowledge. While artificial neural networks excel at learning from stationary datasets, once the i.i.d. assumption is violated, they have been shown to suffer from \textit{catastrophic forgetting} \cite{mccloskey1989catastrophic}, whereby training on new data quickly erases knowledge acquired from older data.

Evaluation of techniques that tackle the catastrophic forgetting problem has largely focused on sequential learning of \textit{distinct} tasks. As such, many successful approaches have relied on the awareness of \textit{task boundaries} for the consolidation of previous knowledge \cite{ruvolo2013ella,kirkpatrick2017overcoming,li2017learning}. This can be problematic because sometimes the data distribution faced by a neural network during training may evolve in a gradual way that can not be discretised easily into separate tasks. In reinforcement learning (RL), for example, the data distribution can change at multiple and unpredictable timescales within the training of a \textit{single} task. This can be due to the fact that (i) states are correlated in time, (ii) the agent's policy, which can greatly affect the distribution of its experiences, continuously evolves during training and/or (iii) the dynamics of the agent's environment are non-stationary.

In this paper, we develop an approach to mitigate catastrophic forgetting in a deep RL setting by consolidating the agent's policy over multiple timescales during training, which we call \textit{policy consolidation} (PC). Rather than relying on task boundaries, consolidation in our model occurs at all times, with the agent's policy being continually distilled into a cascade of hidden networks that evolve over a range of timescales. The hidden networks, in turn, distill knowledge back through the cascade into the policy network in order to ensure that the agent's policy does not deviate too much from where it was previously. 

The PC model is motivated by previous work showing that multiple timescale learning at the \textit{parameter} level helped task-agnostic continual learning in the context of simple reinforcement learning tasks \cite{kaplanis2018continual}. In particular, for the PC model, we adapt the model of \cite{kaplanis2018continual} by incorporating concepts from knowledge distillation \cite{hinton2015distilling,rusu2015policy} and from one of the proximal policy optimisation (PPO) algorithms for RL \cite{schulman2017proximal} in order to implement multi-timescale learning directly at the \textit{policy} level.

We evaluate the PC agent's capability for continual learning by training it on a number of continuous control tasks in three non-stationary settings that differ in how the data distribution changes over time: (i) alternating between two tasks during training, (ii) training on just one task, and (iii) in a multi-agent self-play environment. We find that on average the PC model improves continual learning relative to baselines in all three scenarios.

\section{Preliminaries}
\subsection{Reinforcement Learning}
We chose to evaluate our model in a reinforcement learning setting as it provides a particularly interesting challenge for continual learning, where the distribution of states and actions changes \textit{continuously} during the training of even a single task. 

We consider the standard RL paradigm of an agent interacting with an environment. This is formalised as a Markov decision process $\langle \mathcal{S},\mathcal{A},p_s,r \rangle$, where $\mathcal{S}$ and $ \mathcal{A}$ denote the state and action spaces respectively (both continuous in our experiments), $p_s : \mathcal{S} \times \mathcal{S} \times \mathcal{A} \rightarrow [ 0, \infty )$ defines the probability density function of transitioning to state $s_{t+1} \in \mathcal{S}$ conditioned on the agent taking action $a_t \in \mathcal{A}$ in state $s_t \in \mathcal{S}$. Finally, $r: \mathcal{S} \times \mathcal{A} \rightarrow \mathbb{R}$ defines a function that maps each transition $(s_t,a_t)$ to a scalar reward.

The goal of the RL agent is to determine a policy, defined as a probability density function over actions given state $\pi(a_t | s_t)$, that maximises the expected sum of future rewards:
\begin{equation}
\pi^* = \arg \max_{\pi} \sum_t \mathbb{E}_{\pi} [ r(s_t,a_t) ]
\end{equation}

\subsection{Policy gradient methods}
Policy gradient (PG) methods are a family of reinforcement learning algorithms that aim to maximise the expected return of the agent by directly estimating and ascending its gradient with respect to the parameters that define the agent's policy. The most typical gradient estimators are likelihood ratio policy gradients, which take the form $\mathbb{E}[\hat{A}_t \nabla_\theta \log \pi_\theta]$, where $\pi_\theta$ is the policy defined by parameters $\theta$ and $\hat{A}_t$ is an advantage function. The advantage function typically takes the form of the difference between the realised return of the agent, $\sum_t r(s_t, a_t)$, and a baseline that reduces the variance of the estimator. PG methods have recently been shown to perform well on the types of continuous control tasks that we experiment on in this paper \cite{schulman2017proximal, lillicrap2015continuous}.

\subsection{PPO}
One difficulty with PG methods arises from the fact that the magnitude of a gradient step in parameter space is often not proportional to its magnitude in \textit{policy} space. As such, a small step in parameter space can sometimes cause an excessively large step in policy space, leading to a collapse in performance that is hard to recover from. The Proximal Policy Optimization (PPO) algorithms \cite{schulman2017proximal} tackle this problem by optimising a surrogate objective that penalises changes to the parameters that yield large changes in policy. For example, the objective for the fixed-KL version of PPO is given by:
\begin{align}
\begin{split}
L^{KL} (\theta) = \mathbb{E}_t \Big[&\frac{\pi_\theta (a_t | s_t)}{\pi_{\theta_{old}}(a_t | s_t)}\hat{A}_t - \\ &\beta D_{\mathrm{KL}}\left(\pi_{\theta_{old}}(\cdot | s_t) || \pi_{\theta}(\cdot | s_t)\right)\Big]
\end{split}
\end{align}
where the first term in the expectation comprises the likelihood policy gradient, which uses the generalised advantage estimate for $\hat{A}_t$ \cite{schulman2015high}, and the second term penalises the Kullback-Leibler divergence between the policy and where it was at the start of each update.  At each iteration, trajectories are sampled using $\pi_{\theta_{old}}$ and then used to update $\pi_{\theta}$ over several epochs of stochastic gradient descent of $L^{KL} (\theta)$. The coefficient $\beta$ controls the magnitude of the penalty for large step sizes in policy space. 

We use the various PPO algorithms as baselines in all our experiments. We shall see later on that the policy consolidation model can be viewed as an extension of the fixed-KL version of PPO operating at multiple timescales.

\subsection{Multi-agent RL with Competitive Self-play}

While non-stationarity arises in single agent RL due to correlations in successive states and changes in the agent's policy, the dynamics of the environment given by $p_s(s_{t+1} | s_t, a_t)$ are typically \textit{stable}. In \textit{multi}-agent RL, the environment dynamics are often \textit{unstable} since the observations of one agent are affected by the actions of the other agents, whose policies may also evolve over time. 
In competitive multi-agent environments, training via \textit{self-play}, whereby the agents' policies are all governed by the same controller, has had several recent successes \cite{silver2016mastering,OpenAI_dota}. 

One reason cited for the success of self-play is that agents are provided with the perfect curriculum as they are always competing against an opponent of the same calibre. In practice, however, it has been reported that only training an agent against the most recent version(s) of itself can lead to instabilities during training \cite{heinrich2016deep}. For this reason, agents are normally trained against historical versions of themselves to ensure stability and continuous improvement, often sampling from their entire history \cite{bansal2018emergent}. In a continual learning setting it may not be possible to store all historical agents and the training time will become increasingly prohibitive as the history grows. In this paper, we evaluate the continual learning ability of our model in a self-play setting by only training each agent against the most recent version of itself.

\section{From Synaptic Consolidation to Policy Consolidation}

Recent work has shown that a model of multi-timescale learning at the \textit{synaptic} level \cite{benna2016computational} can alleviate catastrophic forgetting in RL agents \cite{kaplanis2018continual}. One of the theoretical limitations of this model was that, while it improved memory at the level of individual parameters, this would not guarantee improved \textit{behavioural} memory of the agent, due to a highly nonlinear relationship between the parameters and the output of the network. The PC model addresses this limitation by consolidating memory directly at the behavioural level. 

The PC model can also be viewed as an extension of one of the PPO algorithms \cite{schulman2017proximal}, which ensures that the policy does not change too much with every policy gradient step - in a sense, preventing catastrophic forgetting at a very short timescale. The PC agent operates on the same principle, except that its policy is constrained to be close to where it was at several stages in its history, rather than just at the previous step.

Below we motivate and derive the policy consolidation framework: first, we reinterpret the synaptic consolidation model by incorporating it into the objective function of the RL agent, and then we combine it with concepts from PPO and knowledge distillation in order to directly consolidate the agent's behavioural memory at a range of timescales. 

\begin{figure*}[h]
    \begin{subfigure}{0.5 \textwidth}
    \centering
    \includegraphics[width=\textwidth]{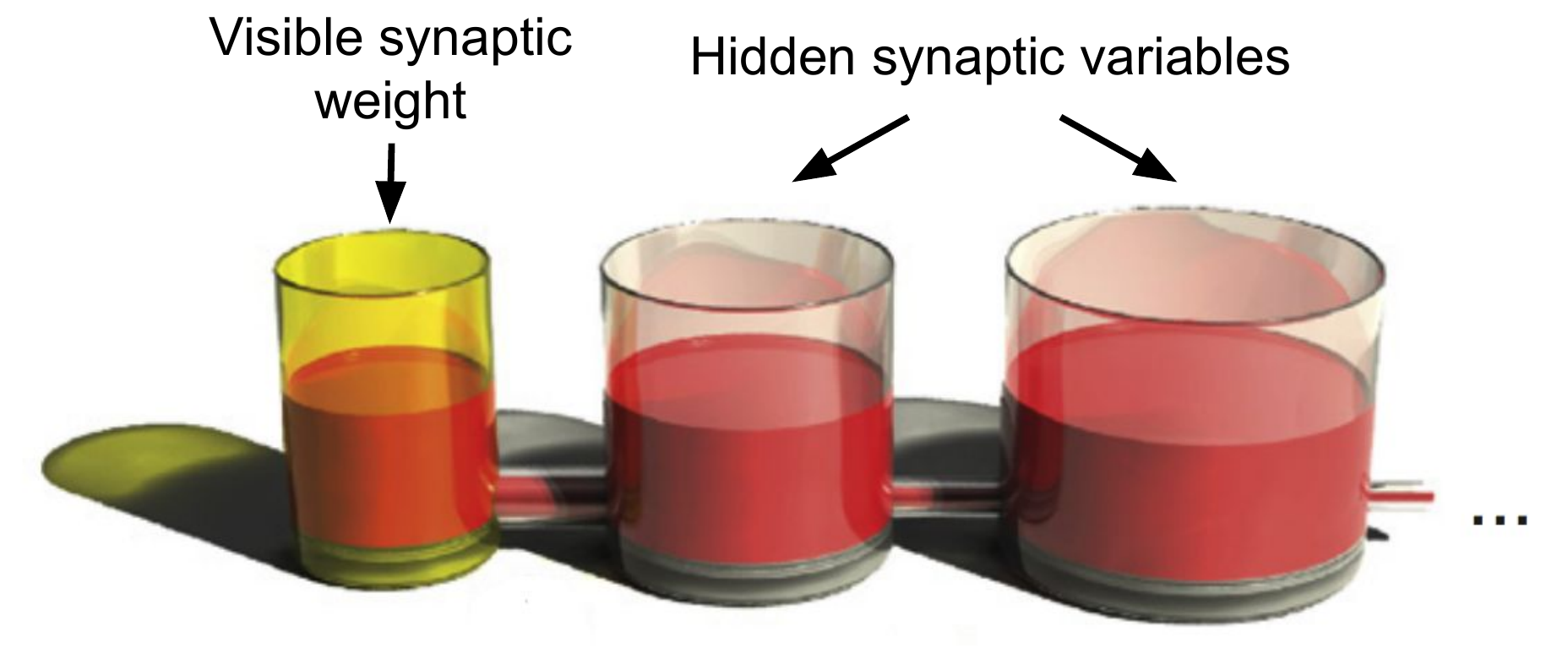} \\
    \includegraphics[width=0.7\textwidth]{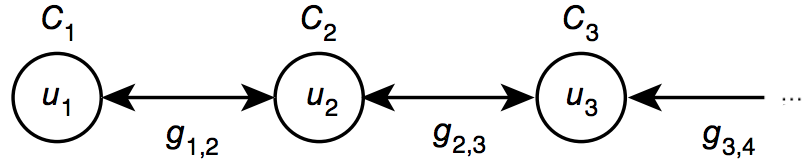}
    \caption{Synaptic Consolidation}
    \label{fig:beakers}
    \end{subfigure}
    \begin{subfigure}{0.5 \textwidth}
    \centering
    \includegraphics[width=\textwidth]{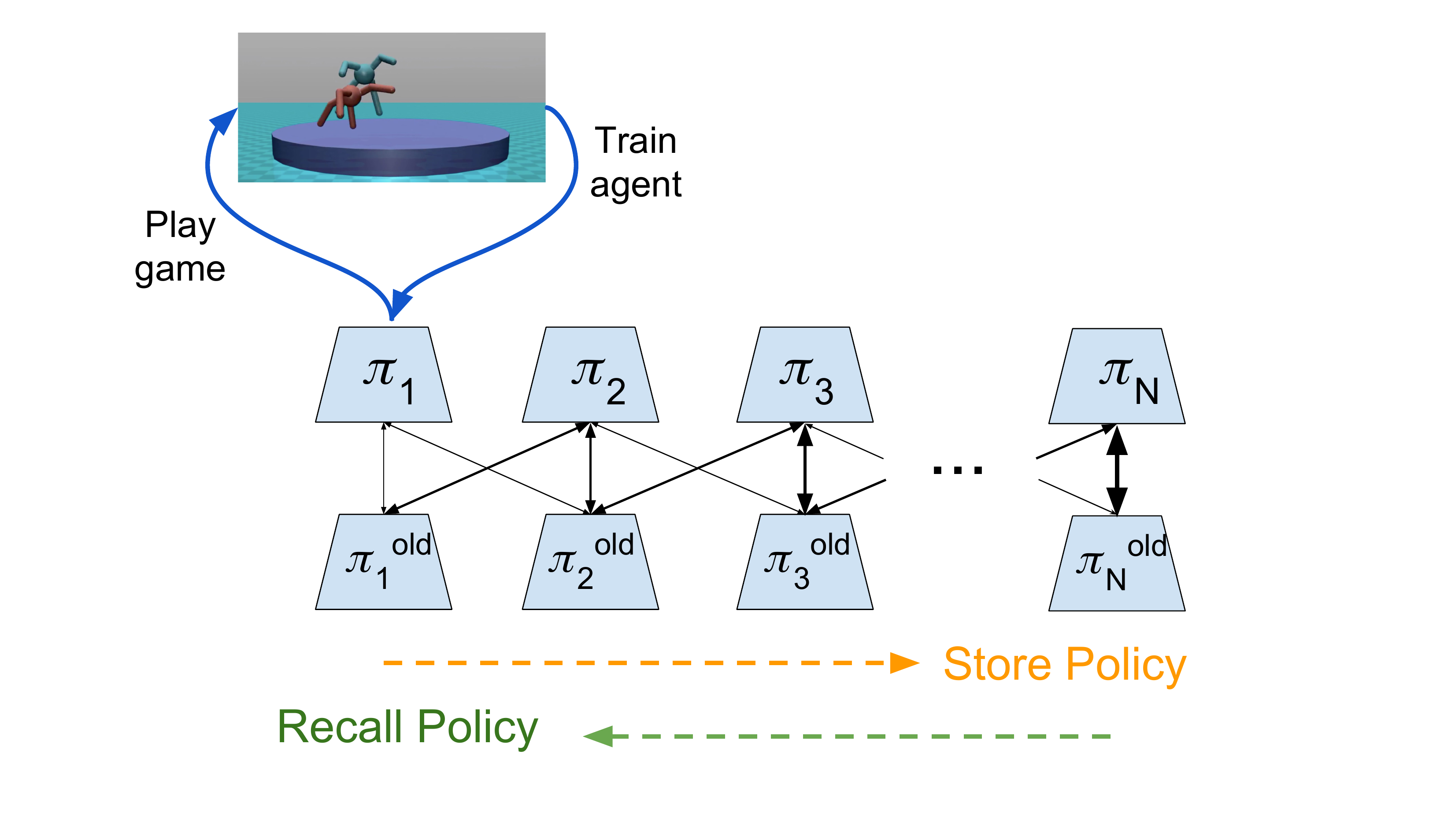}
    \caption{Policy Consolidation}
    \label{fig:kl_beakers}
    \end{subfigure}
    \caption{(a) Depiction of synaptic consolidation model (adapted from \cite{benna2016computational}) (b) Depiction of policy consolidation model. The arrows linking the $\pi_k$s to the $\pi_k^{old}$s represent KL constraints between them, with thicker arrows implying larger constraints, enforcing the policies to be closer together.}
    \vspace{-0.5cm}
\end{figure*}

\subsection{Synaptic Consolidation}
The synaptic model from \cite{benna2016computational} was originally described by analogy to a chain of communicating beakers of liquid (Figure \ref{fig:beakers}). The level of liquid in the first beaker corresponds to the visible synaptic weight, i.e. the one that is used for neural computation. Liquid can be added or subtracted from this beaker in accordance with any synaptic learning algorithm. The remaining beakers in the chain correspond to `hidden' synaptic variables, which have two simultaneous functions: (i) the flow of liquid from shallower to deeper beakers \textit{record} the value of the synaptic weight at a wide range of timescales, and (ii) the flow from deeper beakers back through the shallower ones \textit{regularise} the synaptic weight by its own history, constraining it to be close to previous values. The wide range of timescales is implemented by letting the tube widths between beakers decrease exponentially and the beaker widths to grow exponentially as one traverses deeper into the chain.

The synaptic consolidation process can be formally described with a set of first-order linear differential equations, which can be translated into discrete updates with the Euler method as follows:
\begin{align}
\begin{split}
u_1 &\leftarrow u_1 + \frac{\eta}{C_1}\left(\Delta w +g_{1,2}(u_2-u_1)\right) \\
u_k &\leftarrow u_k + \frac{\eta}{C_k}\left(g_{k-1,k}(u_{k-1}-u_k)+g_{k,k+1}(u_{k+1}-u_k)\right)
\end{split}
\label{eq:bf_1}
\end{align}
where $u_k$ corresonds to the value of the kth variable in the chain (for $k>1$), the $g_{k,k+1}$s correspond to the tube widths, the $C_k$s correspond to the beaker widths, $\Delta w$ corresponds to the learning update and $\eta$ is the learning rate. 

Now consider, as in \cite{kaplanis2018continual}, that we apply this model to the parameters of a neural network that encodes the policy of an RL agent. Let $\textbf{U}_k$ denote a vector of the kth beaker values for all the parameters and let $\mathcal{L}(\textbf{U}_1)$ be the RL objective that the network is being trained to minimise. If we define a new loss function $\mathcal{L}^*(\textbf{U})$ as follows:

\begin{align}
\mathcal{L}^*(\textbf{U})=\mathcal{L}(\textbf{U}_1) +\frac{1}{2} \sum_{k=1}^{N-1} g_{k,k+1}||\textbf{U}_k-\textbf{U}_{k+1}||_2^2
\end{align}

then we notice that, by differentiating it with respect to $\textbf{U}_k$, a negative gradient step with learning rate $\frac{\eta}{C_k}$ implements the consolidation updates in Equation \ref{eq:bf_1} since
\begin{equation}
-\nabla_{\textbf{U}_k} \mathcal{L}^*(\textbf{U}) = g_{k-1,k}(\textbf{U}_{k-1}-\textbf{U}_{k})+g_{k,k+1}(\textbf{U}_{k+1}-\textbf{U}_{k})
\end{equation}
for $k>1$ and a step in the direction of $-\nabla_{\textbf{U}_1} \mathcal{L}(\textbf{U}_1)$ corresponds to $\Delta w$ in Equation \ref{eq:bf_1}. Thus we can view the synaptic consolidation model as a process that minimises the \textit{Euclidean distance} between the vector of parameters and its own history at different timescales. It is not obvious, however, that distance in parameter space is a good measure of \textit{behavioural} dissimilarity of the agent from its past.

\subsection{Policy Consolidation}

Since each parameter has its own chain of hidden variables, each $\textbf{U}_k$ actually defines its own neural network and thus its own policy, which we denote $\pi_k$. With this view, we propose a new loss function that replaces the Euclidean distances in parameter space, given by the $||\textbf{U}_k-\textbf{U}_{k+1}||_2^2$ terms, with a distance in \textit{policy space} between adjacent beakers:
\begin{align}
\begin{split}
\mathcal{L}^{*}(\bm{\pi})=& \mathcal{L}(\pi_1)+ \mathbb{E}_{s_t \sim \rho_1} \Big[ \sum_{k=1}^{N-1} g_{k,k+1}  D_{\mathrm{KL}}\left(\pi_k || \pi_{k+1}\right) \Big]
\end{split}
\end{align}
where $\bm{\pi}=(\pi_1,...,\pi_N)$, $\rho_1$ is the state distribution induced by following policy $\pi_1$ and the  $D_{\mathrm{KL}}\left(\pi_{k} || \pi_{k+1}\right)$ terms refer to the Kullback-Leibler (KL) divergence between the action distributions (given state $s_t$) of adjacent policies in the chain. 

In order to implement policy consolidation in a practical RL algorithm, we adapt the fixed-KL version of PPO \cite{schulman2017proximal}. As a reminder, this version of PPO features a cost term of the form $\beta D_{\mathrm{KL}}\left(\pi_{\theta_{old}}|| \pi_{\theta}\right)$, where $\beta$ controls the size of the penalty for large updates in policy space.

In our model, we use the same policy gradient as in PPO and introduce similar KL terms for each $\pi_k$ with $\beta_k$ coefficients that increase exponentially for deeper beakers, with $\beta_k = \beta \omega^{k-1}$. This ensures that the deeper beakers evolve at longer timescales in \textit{policy} space, with the $\beta_k$ terms corresponding to the $C_k$ terms in the synaptic consolidation model. We found in our experiments that the PC agent's performance was often more stable when we used the reverse KL constraint $D_{\mathrm{KL}}\left( \pi_k || \pi_{k_{old}}\right)$ (Appendix B), and so the final objective for the PC model was given by:
\begin{align}
L^{PC}(\bm{\pi}) = L^{PG}(\pi_1)+L^{PPO}(\bm{\pi})+L^{CASC}(\bm{\pi})
\end{align}
where $L^{PG}(\pi_1)$ is the policy gradient $\mathbb{E}_t \Bigg[\frac{\pi_1}{\pi_{1_{old}}}\hat{A}_t \Bigg]$,
$L^{PPO}(\bm{\pi})$ constitutes the PPO constraints that determine the timescales of the $\pi_k$s,
\begin{align}
L^{PPO}(\bm{\pi}) = -\mathbb{E}_t \Bigg[\sum_{k=1}^{N} \Big[\beta \omega^{k-1} D_{\mathrm{KL}}\left(\pi_k || \pi_{k_{old}}\right)\Big]\Bigg]
\end{align}
and $L^{CASC}(\bm{\pi})$ captures the KL terms between neighbouring policies in the cascade,
\begin{align}
\begin{split}
L^{CASC}(\bm{\pi}) = -\mathbb{E}_t \Bigg[\omega_{1,2} D_{\mathrm{KL}}\left(\pi_1\right || \pi_{2_{old}})+ \\ \sum_{k=2}^{N} \Big[\omega D_{\mathrm{KL}}\left(\pi_k\right || \pi_{{k-1}_{old}})+D_{\mathrm{KL}}\left(\pi_k\right || \pi_{{k+1}_{old}})\Big]\Bigg]
\end{split}
\end{align}

where $\pi_{N+1_{old}}:=\pi_N$, $\omega_{1,2}$ controls how much the agent's policy is regularised by its history and $\omega > 1$ determines the ratio of timescales of consecutive beakers. A smaller $\omega$ gives a higher granularity of timescales along the cascade of networks, but also a smaller maximum timescale determined by $\beta_N = \beta \omega^{N-1}$. Each policy is constrained to be close to the \textit{old} versions of neighbouring policies in order to ensure a stable optimisation at each iteration (Figure \ref{fig:kl_beakers}).


\section{Experiments}

In order to test how effective the model is at alleviating catastrophic forgetting, we evaluated the performance of a PC agent over a number of continuous control tasks \cite{brockman2016openai,henderson2017multitask, al-shedivat2018continuous} in three separate RL settings. The settings are differentiated by the nature of how the data distribution changes over time in each case:
\begin{enumerate}[(i)]
    \item In the first setting, the agent was trained alternately on pairs of separate tasks. This is akin to the most common continual learning setting in the literature, where the changes to the distribution are \textit{discrete}. While in other methods, these discrete transitions are often used to inform consolidation in the model \cite{kirkpatrick2017overcoming,zenke2017continual}, in the PC model the consolidation process has no explicit knowledge of task transitions.
    \item In the second setting, the agent was trained on single tasks for long uninterrupted periods. The goal here was to test how well the PC model could handle the \textit{continual} changes to the distribution of experiences caused by the evolution of the agent's policy during training. Policy-driven changes to the state distribution have previously been shown to cause instability and catastrophic forgetting in continuous control tasks \cite{de2016off, kaplanis2018continual}. In this case the dynamics of the environment, given by $p_s(s_{t+1} | s_t, a_t)$, are stationary.
    \item In the third setting, the agent was trained in a competitive two-player environment via self-play, whereby the same controller was used and updated for the policy of each player. In this case, the distribution of experiences of each agent are not only affected by its own actions, but also, unlike the single agent experiments, by changes to the state transition function that are influenced by the opponent's evolving policy.
\end{enumerate}
 In each setting, baseline agents were also trained using the fixed-KL, adaptive-KL and clipped versions of PPO \cite{schulman2017proximal,baselines} with a range of $\beta$s and clip coefficients for comparison. The results presented in the main paper use the \textit{reverse} KL constraint for the fixed- and adaptive-KL baselines since that was what was used for the PC agent, but we observed similar results using the original PPO objectives (Appendix B).
 
 \subsection{Single agent experiments}
 
 
 \begin{figure*}[h]
    \begin{subfigure}{0.5 \textwidth}
    \centering
    \includegraphics[width=\textwidth,height=4.3cm]{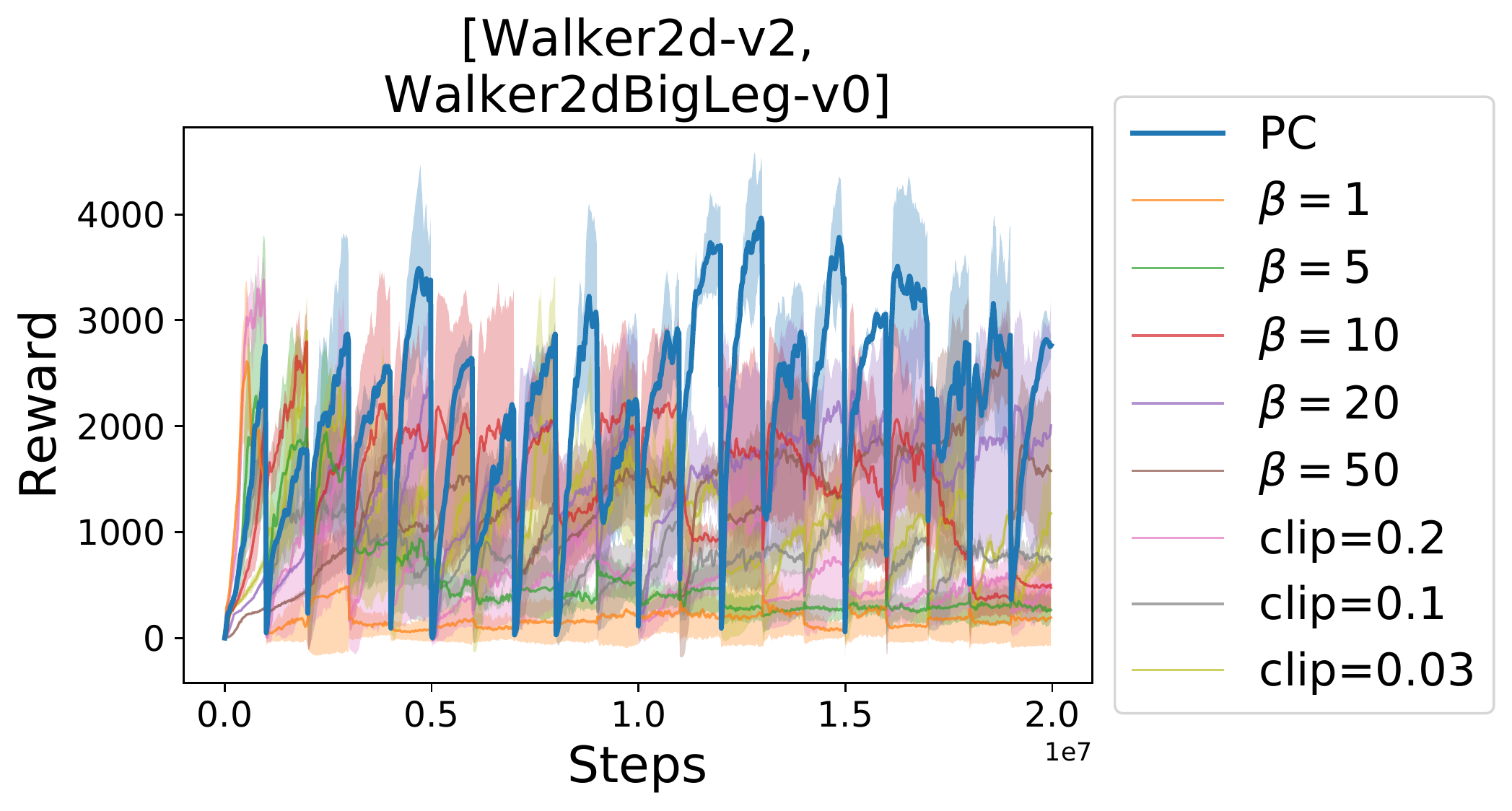}
    \end{subfigure}
    \begin{subfigure}{0.5 \textwidth}
    \centering
    \includegraphics[width=\textwidth,height=4.3cm]{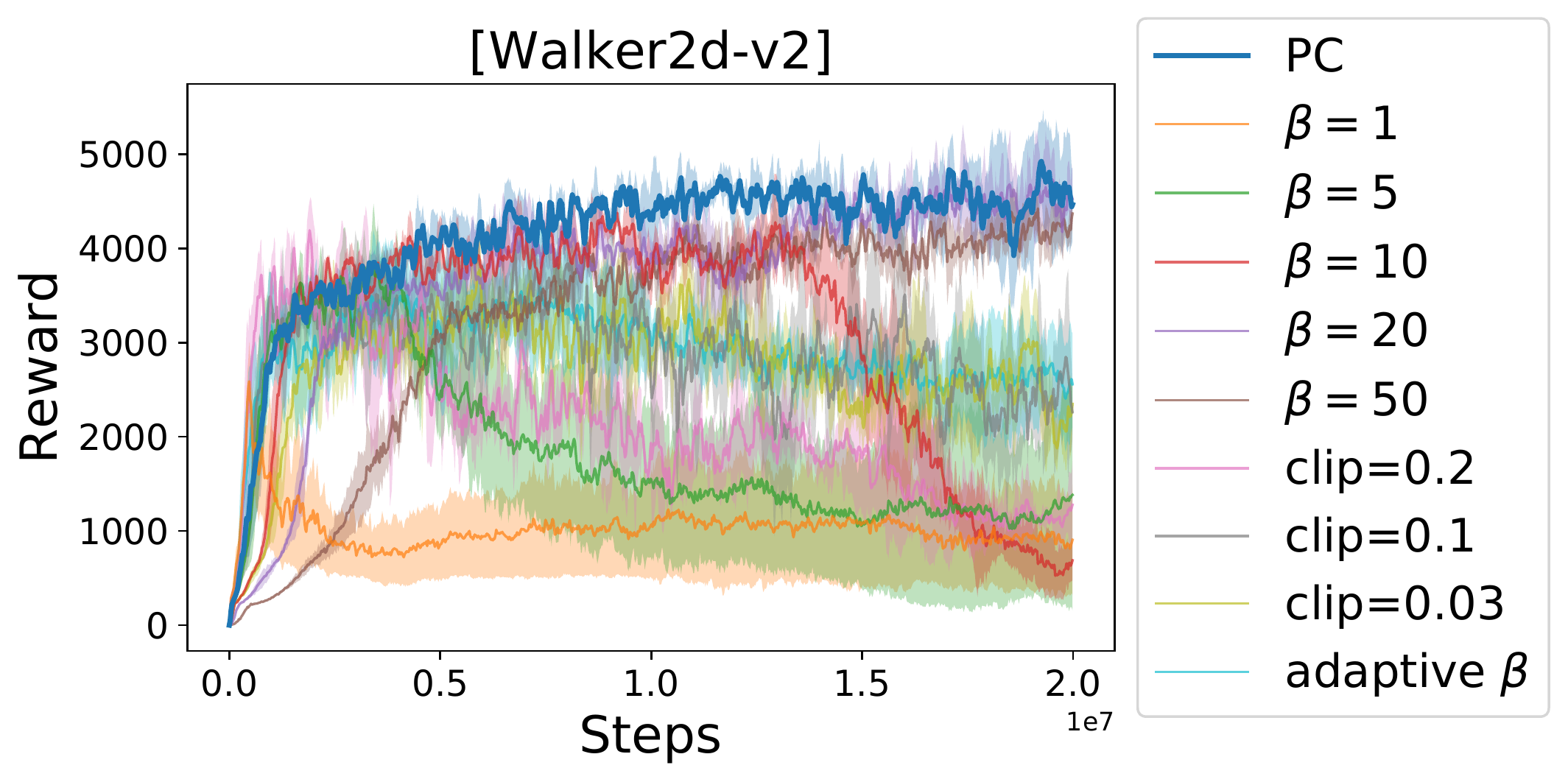}
    \end{subfigure}
    \begin{subfigure}{0.5 \textwidth}
    \centering
    \includegraphics[width=\textwidth,height=4.3cm]{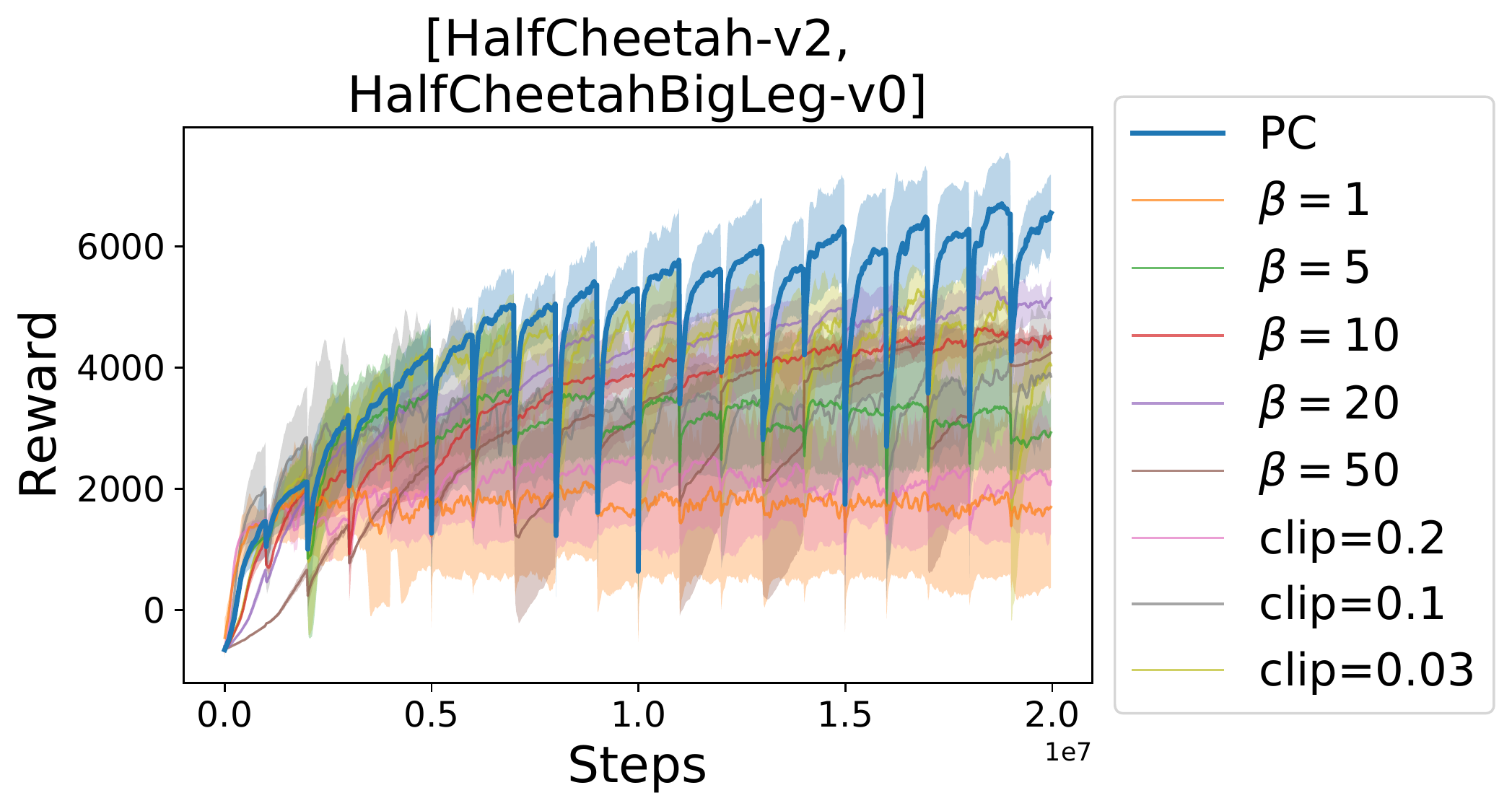}
    \end{subfigure}
    \begin{subfigure}{0.5 \textwidth}
    \centering
    \includegraphics[width=\textwidth,height=4.3cm]{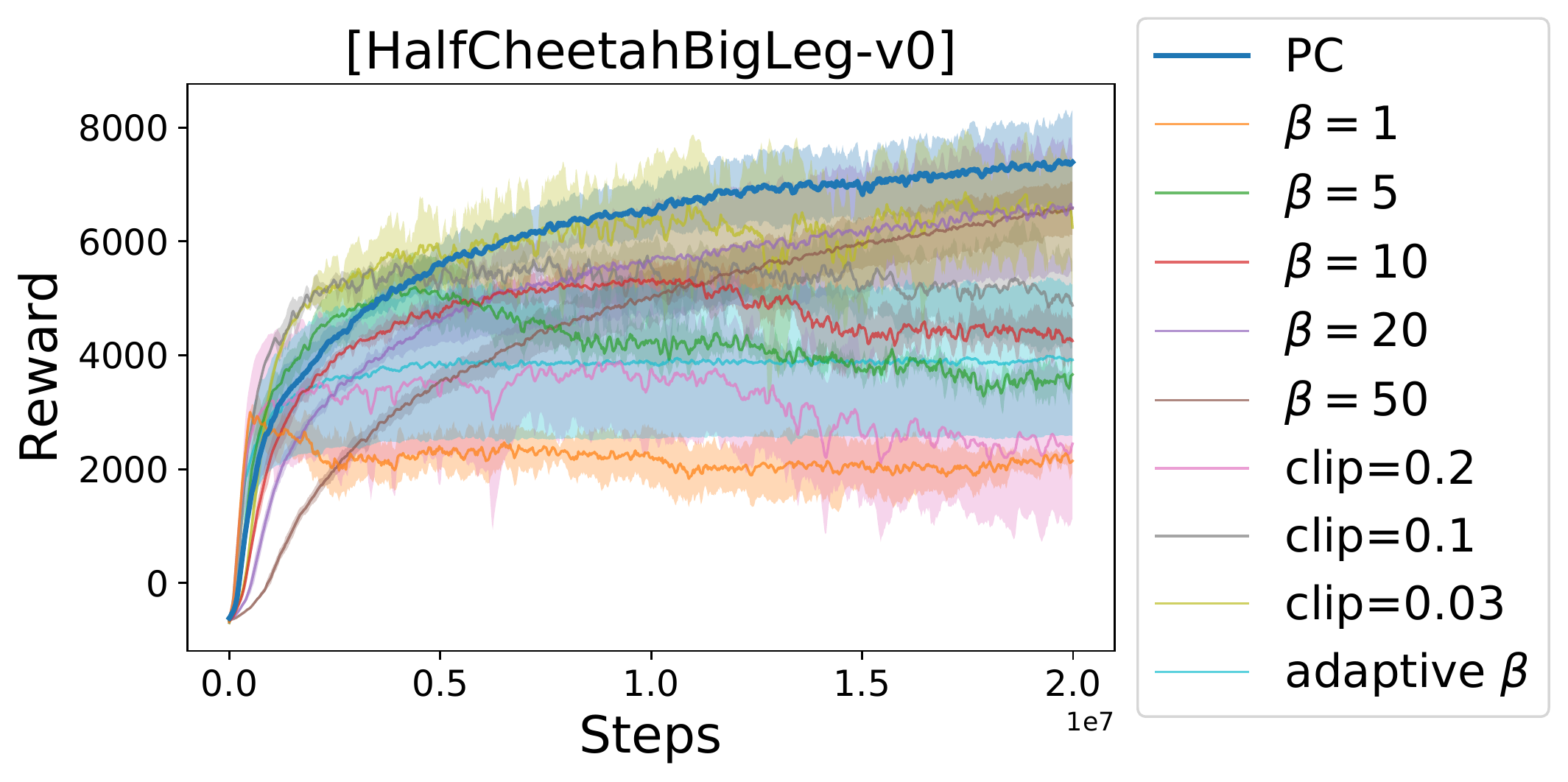}
    \end{subfigure}
    \begin{subfigure}{0.5 \textwidth}
    \centering
    \includegraphics[width=\textwidth,height=4.3cm]{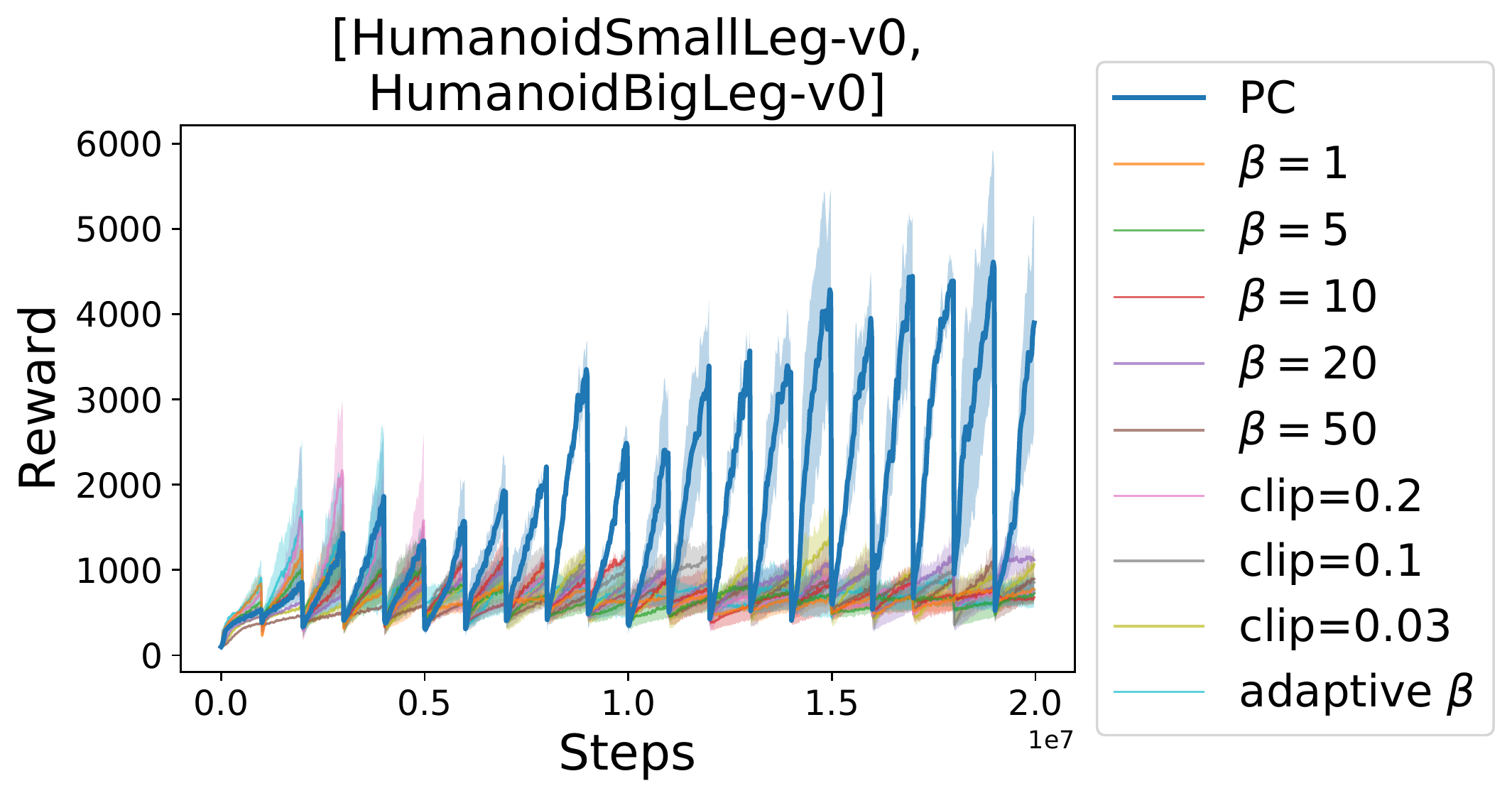}
    \caption{Alternating tasks}
    \label{fig:twotask}
    \end{subfigure}
    \begin{subfigure}{0.5 \textwidth}
    \centering
    \includegraphics[width=\textwidth,height=4.3cm]{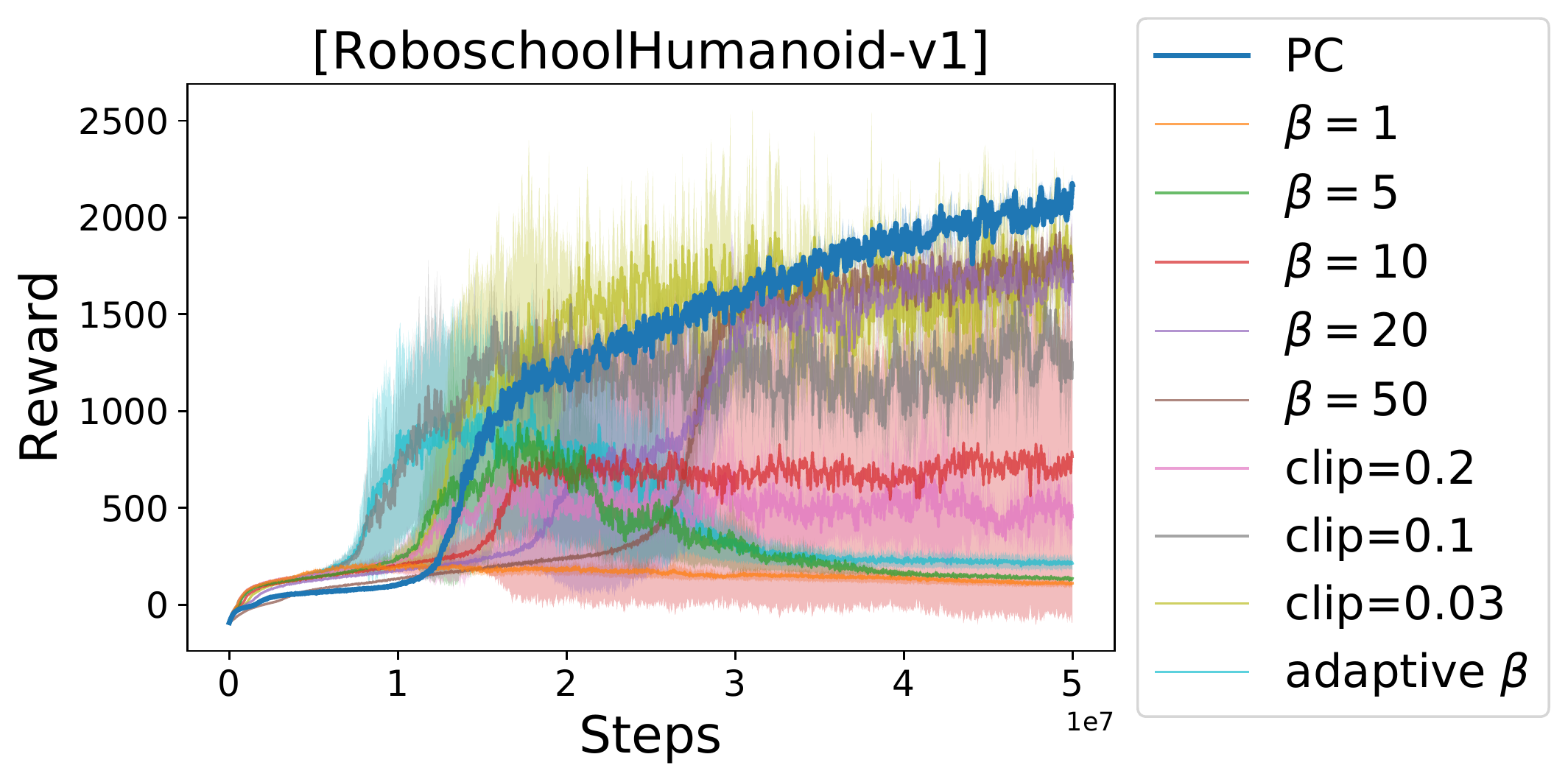}
    \caption{Single task}
    \label{fig:onetask}
    \end{subfigure}
    \caption{Reward over time for (a) alternating task and (b) single task runs; comparison of PC agent with fixed KL (with different $\beta$s), clipped (with different clip coefficients) and adaptive KL agents (omitted for some runs since return went very negative). Means and s.d. error bars over 3 runs per setting.}
    \vspace{-0.5cm}
\end{figure*}
 
 \subsubsection{Setup}
 All agents shared the same architecture for the policy network, namely a multilayer perceptron with two hidden layers of 64 ReLUs. The PC agent used for all experiments (unless otherwise stated) consisted of 7 hidden policy networks with $\beta=0.5$ and $\omega=4.0$. The hyperparameters used for training were largely the same as those used for the Mujoco tasks in the the original PPO paper \cite{schulman2017proximal}. The hyperparameters were also constant across tasks, except for the learning rate, which was lower for all agents (including the baseslines) in the Humanoid tasks.

In the alternating task setting, the task was switched every 1 million time steps for a total of 20 million steps of training. Three pairs of continuous control tasks were used (taken from \cite{brockman2016openai,henderson2017multitask, schulman2017proximal}); the tasks in each pair were similar to one another, but different enough to induce forgetting in the baseline agents in most cases. In the single task setting, agents were trained for either 20mn or 50mn time steps per task. In both settings, the mean reward on the current task was recorded during training. Full implementational details are included in Appendix A1.

\subsubsection{Results}

In the alternating task setting, the PC agent performs better on average than any of the baselines, showing a particularly stark improvement in the Humanoid tasks where none of the baselines were able to successfully learn both tasks. The PC agent, on the other hand, was able to continue to increase its mean reward on each task throughout training (Figure \ref{fig:twotask}). In the single task setting, the PC agent does as well as or better than the baselines in all tasks and also exhibits strikingly low variance in the Humanoid task (Figure \ref{fig:onetask}).

\subsection{Multi-agent experiments}

\subsubsection{Setup}
For the multi-agent experiments, agents were trained via self-play in the RoboSumo-Ant-vs-Ant-v0 environment developed in \cite{al-shedivat2018continuous}. 
The architecture of the agents in the self-play experiments was the same as in the single-agent runs, but some of the hyperparameters for training were altered, mainly due to the fact that training required many more time steps of experience than in the single agent runs. One important change was that the batch sizes per update were much larger as the trajectories were made longer and also generated by multiple distributed actors (as in \cite{schulman2017proximal}). A larger batch size reduces the variance of the policy gradient, which allowed us to permit larger updates in policy space by decreasing $\beta$ and  $\omega_{1,2}$ (from 0.5 and 4 to 0.1 and 0.25 respectively) in the PC model and thus speed up training. Full implementational details given in Appendix A2.

Whilst we did not train the agents on past versions of themselves, we saved the historical models for evaluation purposes at test time. During training, episodes were allowed to have a maximum length of 500 time steps, but this was increased to 5000 at test time in order to reduce the number of draws between agents of similar ability and more easily differentiate between them.

\subsubsection{Results}
\begin{figure*}[h]
    \centering
    \begin{subfigure}{0.45 \textwidth}
    \centering
    \includegraphics[width=\textwidth]{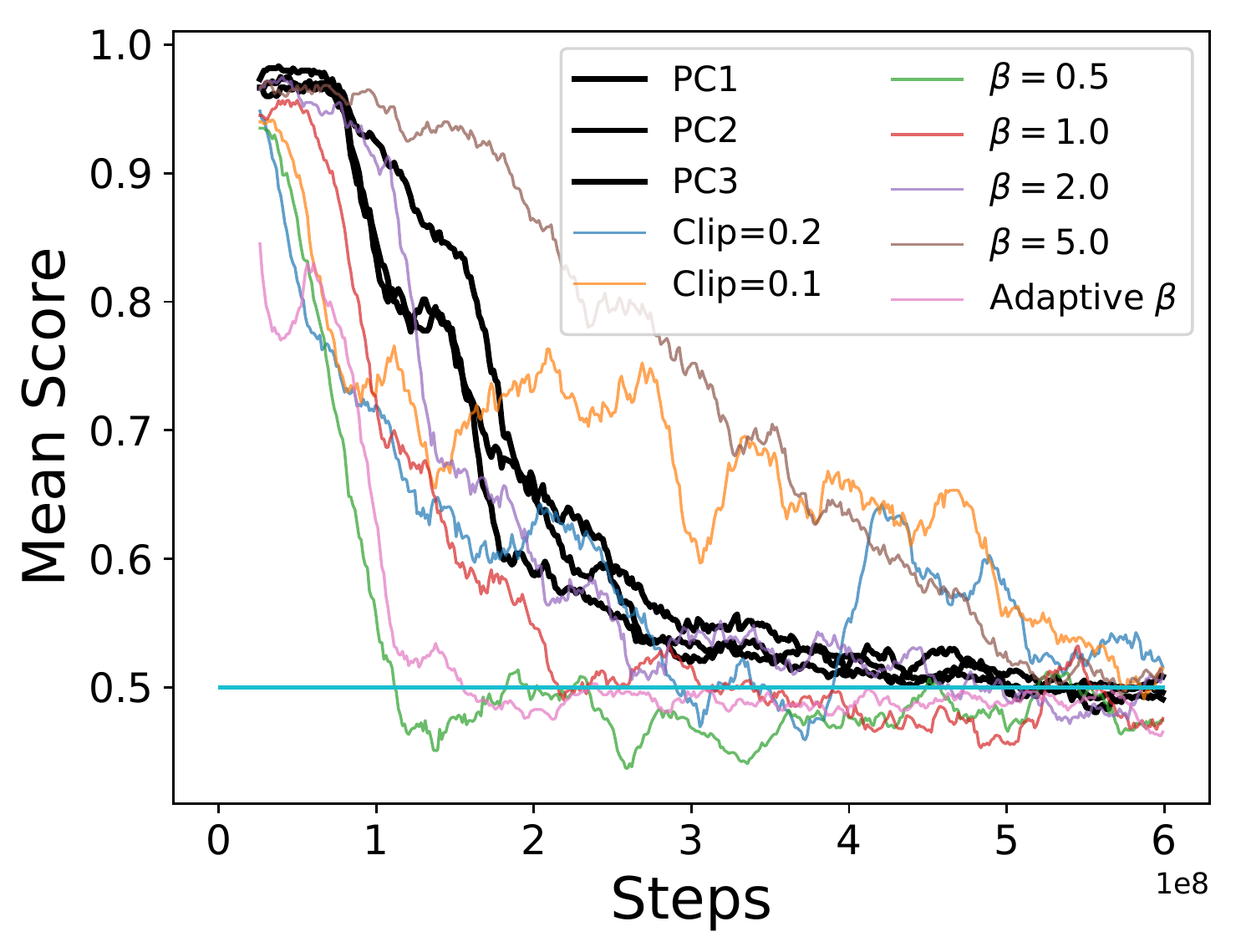}
    \caption{Final model vs. self history}
    \end{subfigure}
    \begin{subfigure}{0.45 \textwidth}
    \centering
    \includegraphics[width=\textwidth]{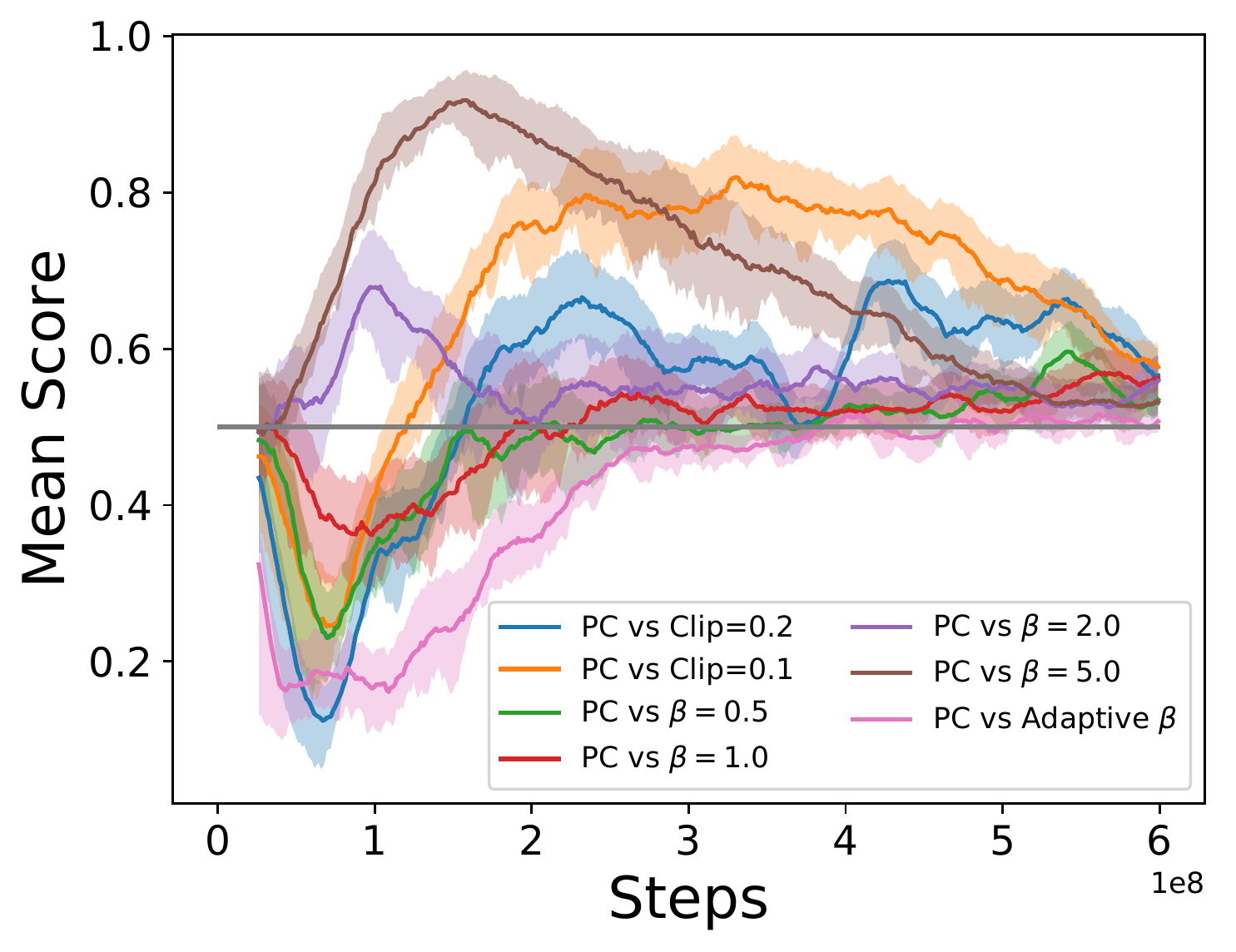}
    \caption{PC vs. baselines over training}
    \end{subfigure}
    \caption{Moving averages of mean scores over time in RoboSumo environment of (a) the final version of each model against its past self at different stages of its history, and (b) the PC agents against the baselines at equivalent points in history. Mean scores calculated over 30 runs using 1 for a win, 0.5 for a draw and 0 for a loss. Error bars in (b) are s.d. across three PC runs, which are shown individually in (a).}
    \label{fig:selfplay}
    \vspace{-0.5cm}
\end{figure*}

 The first experiment we ran post-training was to pit the final agent for each model against the past versions of itself at the various stages of training. The final PC agents were better than almost all their historical selves, only being marginally beaten by a few of the agents very late on in training. Additionally, the decline in performance against later agents was relatively monotonic for the PC agents, indicating a smooth improvement in the capability of the agent. The fixed-KL agents with low $\beta$ ($0.5$ and $1.0$) and the adaptive-KL agents were defeated by a significant portion of their historical selves, whilst the clipped-PPO agents exhibited substantial volatility in their performance, demonstrating signs of catastrophic forgetting during training (Figure \ref{fig:selfplay}a).
 
 The second experiment we ran was to match the PC agents against each of the baselines at their equivalent stages of training. While some of the baseline agents were better than the PC agents during the early stages of training, the PC agents were better on average than all the baselines by the end (Figure \ref{fig:selfplay}b). The superiority of the PC agents over the adaptive-KL agent was only marginal but it seemed to be slowly increasing over time and, as mentioned in the previous paragraph, the adaptive-KL agent was showing signs of catastrophic forgetting by the end of training. The two baseline agents that did not appear to exhibit much forgetting in the first experiment, $\beta=2.0,5.0$, were inferior to the PC agents at all stages of training. In the future it would be interesting to train agents for longer to see how their relative performances evolve. Furthermore, it is possible that the PC model is slow to learn at the beginning of training because it is over-consolidated at the initial policy; it would be interesting to implement incremental flow from deeper policies into shallower ones in the PC model as training progresses (as in \cite{kaplanis2018continual}) to see if this problem is resolved.

\subsection{Further Analyses}
\subsubsection{Testing policies of hidden networks}
\begin{figure*}[h]
    \centering
    \begin{subfigure}{0.49 \textwidth}
    \centering
    \includegraphics[width=\textwidth]{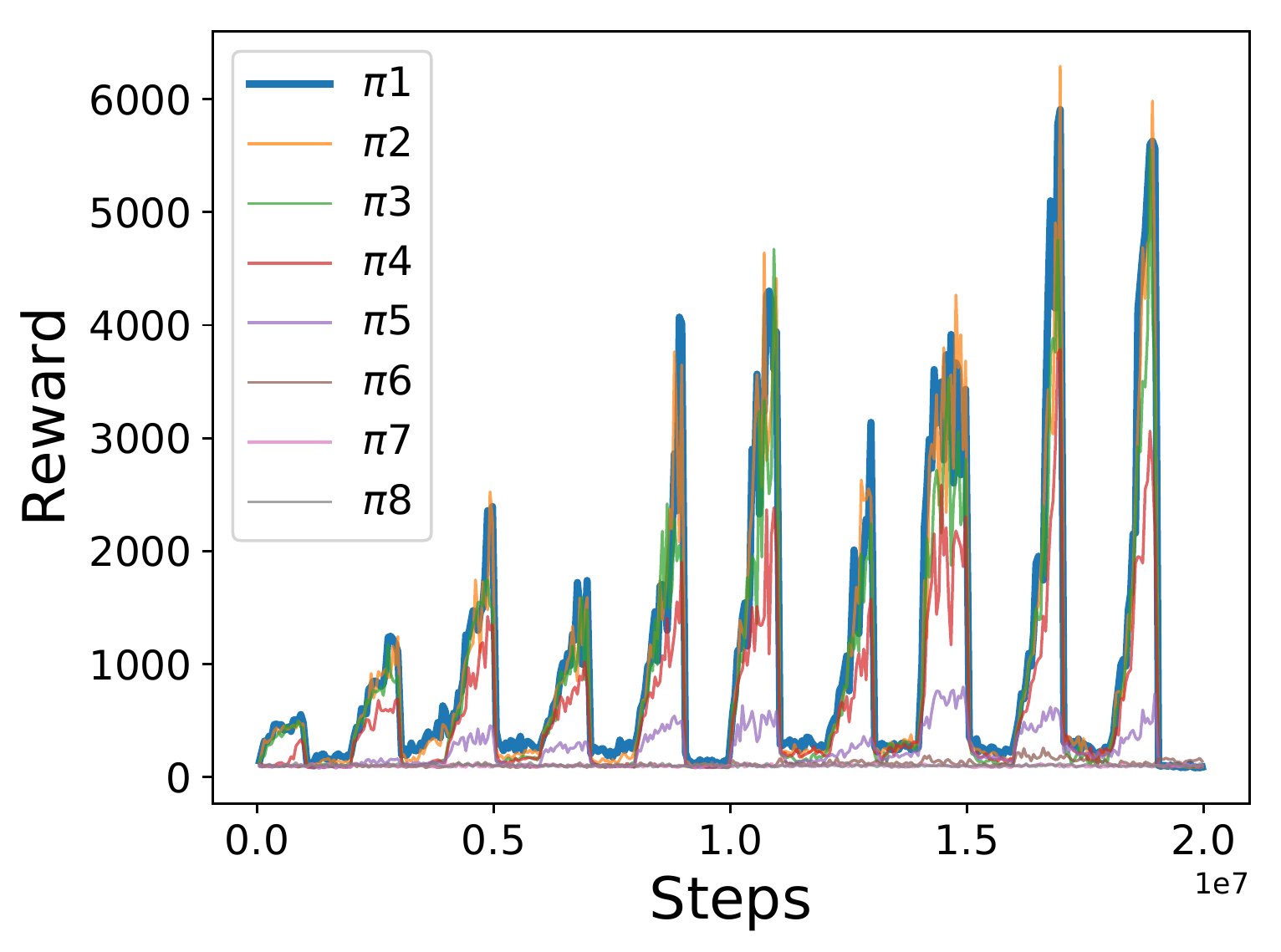}
    \caption{Performance of visible and hidden policies}
    \end{subfigure}
    \begin{subfigure}{0.49 \textwidth}
    \centering
    \includegraphics[width=\textwidth]{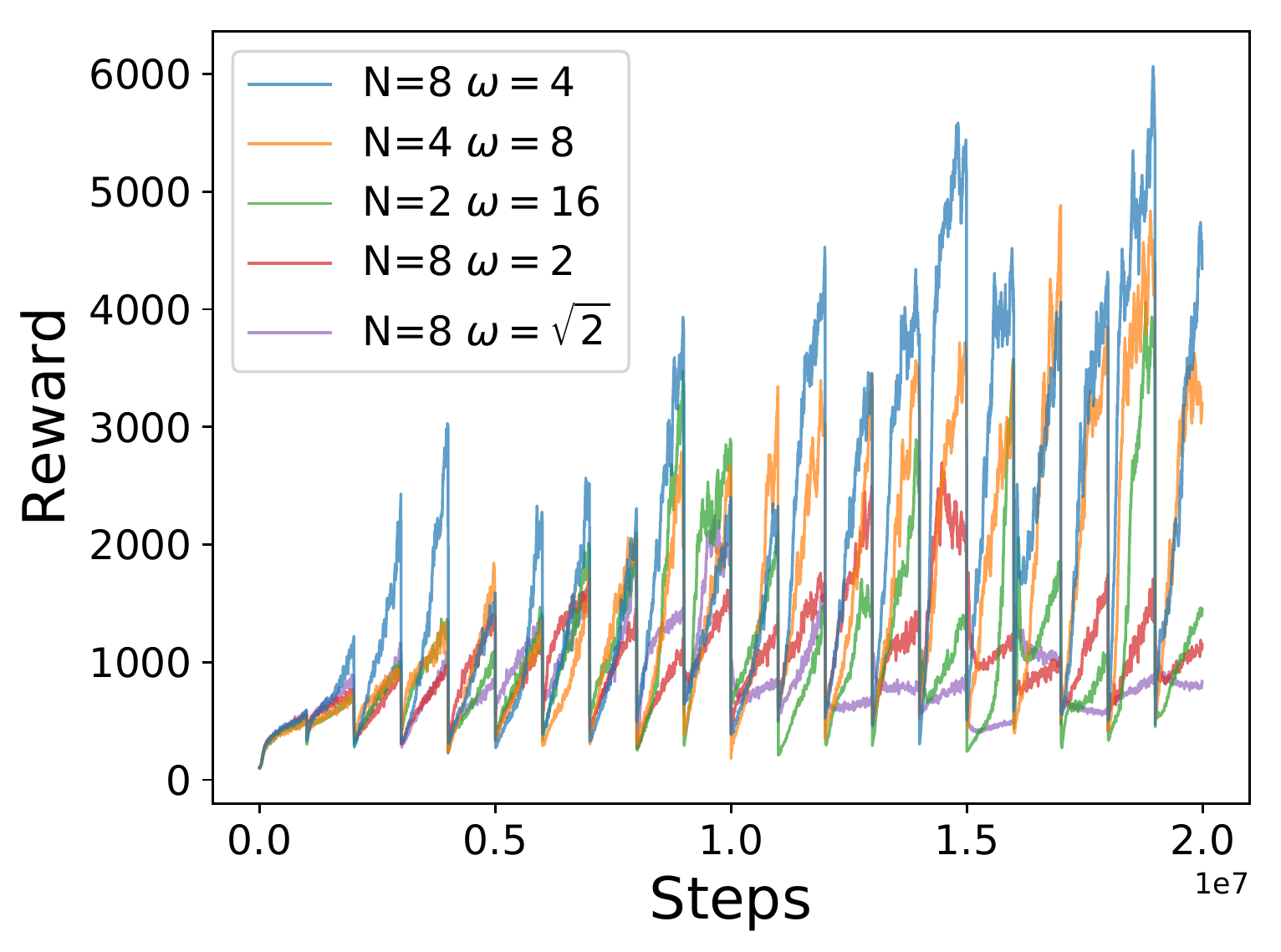}
    \caption{Changing cascade length and granularity}
    \end{subfigure}
    \caption{(a) Reward over time of the policies of the networks at different cascade depths on HumanoidSmallLeg-v0, having been trained alternately on HumanoidSmallLeg-v0 and HumanoidBigLeg-v0. (b) Reward over time on alternating Humanoid tasks for different combinations of cascade length and $\omega$.}
    \label{fig:further}
    \vspace{-0.5cm}
\end{figure*}

In all experiments described thus far, the actions taken by the PC agent were all determined by the policy of the first network in the cascade. However, we thought that testing the performance of the \textit{hidden} policies might provide some insight into the workings of the PC model. 
We evaluated the cascade policies of a PC agent that had been trained on the alternating Humanoid tasks by testing their performance on just one of the tasks at the various stages of training. We observed that (i) all policies quickly drop in performance each time training on the other task commences, and (ii) the visible policy outperforms all the hidden policies at almost all times during training (Figure \ref{fig:further}a). 

It is to be expected that the shallower policies in the cascade will forget quickly when the task is switched, since they operate at short timescales, but one might have expected the deeper policies to maintain a good performance even after a task switch. The results tell us that the memory of the hidden policy networks is not always stored in the form of coherent policies. This phenomenon might be understood by considering how the PC model is trained. 

Currently, the PC model tries to minimise the KL divergence between the conditional one-step action distributions given state between adjacent policies in the cascade; this is \textit{not}, however, the same as minimising the KL divergence between the trajectory distributions of adjacent policies. If we denote the probability of a trajectory $\tau$ while following policy $\pi_k$ as $p_{\pi_k}(\tau)$ then, when using trajectories from $\pi_1$ to estimate $D_{\mathrm{KL}}\left( p_{\pi_k} || p_{\pi_{k+1_{old}}}\right)$ for $k>1$, importance sampling factors must be introduced of the form $\frac{p_{\pi_k}}{p_{\pi_1}}$. In some initial experiments, we found that these factors, which involve large products in the numerator and denominator, introduced huge variance to the updates, especially for deeper policies in the cascade. In the future, it would be interesting to see if this could be done with lower variance and whether it can positively affect the coherence of the hidden policies. This could potentially improve the continual learning ability of the agent and also possibly make it beneficial for the agent to sample from the hidden policies to improve its performance after a task switch.

\subsubsection{Effects of cascade length and granularity}

Further experiments were run in the alternating Humanoid task setting in order to evaluate the importance of (i) the granularity of timescales and (ii) the maximum timescale of the cascade (controlled by $\beta_k$ coefficient of deepest hidden policy) for the continual learning ability of the PC model.
To test the effect of reducing the granularity, we reduced the length of the cascade to 4 and 2 networks (from 8) while maintaining the maximum timescale of the deepest policy ($\beta_N = \beta \times \omega^{N-1}$) by increasing $\omega$ accordingly. To test the effect of decreasing the maximum timescale, we reduced $\omega$ from $4$ to $2$ and $\sqrt{2}$ while keeping the length of the cascade constant at $8$, effectively decreasing $\beta_N$. We found that the original agent with $N=8$ and $\omega=4$ was the best at continual learning, and that the effect of reducing the maximum timescale was much more drastic than that of coarsening the range of timescales (Figure \ref{fig:further}b). 

\subsubsection{Effects of changing task switching schedule}
In order to test that the PC model was robust to different timescales of switching between tasks, we ran experiments on the alternating Humanoid tasks with different task switching schedules. The task switching schedule was altered by factors that differed to the the ratio of timescales between hidden policies to ensure that the continual learning ability of the agent was not the result of a harmonic resonance effect. We found that, while continual learning was still possible with slower schedules, at a much faster schedule the agent struggled to switch between tasks (Appendix C). At first glance, this is perhaps counterintuitive since the fast switching schedule should be closer to the i.i.d. setting and thus easier to learn. However, it could just be that the policies of the two tasks cannot be simultaneously represented easily in the same network. This may be remedied with the introduction of a task-id or a recurrent model that can recognise the task at hand. A more speculative alternative explanation could be that consolidation in the PC model is more effective when learning in blocks, as is thought to be the case for humans learning complex motor skills \cite{wulf2002principles}. Finally, for runs using one baseline, we found that continual learning was not possible for any of the tested schedules (Appendix C).

\section{Related Work}

Approaches to mitigating catastrophic forgetting in neural networks include ones that selectively regularise the parameters to preserve knowledge from previously trained tasks \cite{kirkpatrick2017overcoming, zenke2017continual,li2017learning}, ones that allocate more neural resources over time \cite{ring1997child,rusu2016progressive}, and ones that explicitly store (or train generative models to mimic) past data which are used in various ways to improve the memory of the network \cite{isele2018selective,sprechmann2018memory,shin2017continual}.

The regularisation methods typically use task boundaries as consolidation posts. While in \cite{kirkpatrick2017overcoming}, for example, a method is employed for \textit{detecting} these boundaries if they exist \cite{milan2016forget}, it is still not practical for situations where the data distribution is changing continuously over time. In \cite{kaplanis2018continual}, task boundaries are not required but, as discussed already, the consolidation does not take into account the output or loss function of the network.

Of the methods that grow the amount of neural resources, some utilise the definition of task boundaries \cite{rusu2016progressive, xu2018reinforced}, while others do not \cite{ring1997child, zhou2012online}, but since the computational complexity grows with the size of the network, it is not clear that they are scalable for continual learning. 

The episodic memory approaches for continual learning often do not rely on task boundaries. However, for a fixed memory size, they face the challenge of choosing what data to store over time. Recently it has been shown that simply maintaining a uniform distribution of the data over the lifetime of the model yields decent results \cite{isele2018selective, rolnick2018experience}. Other methods use episodic memory to adapt the model at test time \cite{sprechmann2018memory} or to only allow updates that do not increase the loss on stored examples \cite{lopez2017gradient}. The models that use generative models to mimic older parts of the data distribution suffer from the fact that the generative model can also experience catastrophic forgetting \cite{shin2017continual}.

Distillation of knowledge from one network to another \cite{hinton2015distilling}, a technique we employed for the PC model, has also featured in other continual learning approaches. In \cite{pmlr-v80-schwarz18a}, knowledge is distilled unidirectionally from a flexible network to a more stable one, and vice versa in \cite{furlanello2016active}. The PC model differs from these two in that the distillation is bidirectional and that networks at multiple timescales are used, rather than just two. Bidirectional distillation is also employed in \cite{teh2017distral}, but for transfer learning in a multitask context.

\section{Conclusion and Future Work}
In this paper, we introduced the PC model, which was shown to reduce catastrophic forgetting in a number of RL settings without prior knowledge of the frequency or timing of changes to the agent's distribution of experiences, whether discrete or continuous. 

There are several potential avenues for future work and improvement of the model. A first step would be to run more experiments to compare the PC model to other continual learning methods that are theoretically able to handle continuously changing environments, e.g. \cite{kaplanis2018continual, isele2018selective}, to test the model on a greater variety and number of tasks in sequence, and to perform a more thorough analysis of the hyperparameters of the model. 

One limitation of the PC model is that the action distributions are consolidated equally for every state, while it may be more effective to \textit{prioritise} the storage of particularly important experiences. One way could be to use importance sampling factors, as discussed earlier on, to distill the trajectory distributions rather than single-step action distributions between hidden policies. Another way could be to prioritise consolidation in states where there is large variability in estimated value for the available actions (i.e. ones where the action chosen is particularly crucial).

It is well-known in psychology that the spacing of repetition is important for the consolidation of knowledge in humans \cite{anderson2000learning}. In this vein, it would also be interesting to see if the PC method could be adapted for off-policy RL in order to investigate any potential synergies with experience replay, perhaps incorporating ideas from some of the existing episodic memory methods for continual learning described in the previous section. 

\section*{Acknowledgements}
We would like to thank Matthew Crosby for a thorough critique of the manuscript and Benjamin Beyret for invaluable technical support. This work was supported by BBSRC BB/N013956/1 \& BB/N019008/1, Wellcome Trust 200790/Z/16/Z, Simons Foundation 564408, EPSRC EP/R035806/1, NIH 1R01NS109994-01. 
\bibliography{example_paper}
\bibliographystyle{icml2019}

\title{Supplementary Material}
\date{}
\beginsupplement
\appendix

\maketitle
\section{Details of Implementation}\label{app:imp}
Much of the code for the PC model was built on top of and adapted from the distributed PPO implementation in \cite{baselines}.

\subsection{Single agent experiments}

For the baseline models, we mainly used the hyperparameters used for the training of Mujoco tasks in \cite{schulman2017proximal}. The value function network shared parameters with the policy network and no task-id input was given to the agents. As in \cite{baselines}, the running mean and variance of the inputs was recorded and used to normalise the input to mean 0 and variance 1. The gradients are also clipped to a norm of 0.5 as in \cite{baselines}. In \cite{schulman2017proximal}, different parameters were used for the Humanoid tasks as well as multiple actors - for simplicity we used the Mujoco parameters and a single actor. The hidden policies were all initialised with the same parameters as the visible policy for the PC agent, which means that the beginning of training can be slow as the agent is over-consolidated at the initial weights. This might be remedied in the future by introducing incremental flow from the deeper beakers as training progresses. 

Table \ref{table:params} shows a list of hyperparameters used for the experiments. In future, we would like to do a broader parameter search for both the baselines and the policy consolidation model. For this work, many more baselines were run than policy consolidation agents in the interest of fairness. 

\subsection{Self-play experiments}

For the self-play experiments, the agents were trained for much longer than in the single agent tasks. For this reason, in order to speed up training, a number of changes were made, namely: using multiple environments in parallel to generate experience, increasing the trajectory length, increasing the minibatch size, reducing number of epochs per update. As a result of increasing the number of experiences trained on per update as well as the trajectory length, it was reasonable to expect that the variance of the updates should decrease and that short term non-stationarity is better dealt with. For this reason, we reduced $\omega_{1,2}$ and $\beta$ in the PC model to allow larger updates per iteration. Additionally, we compared the PC model to a lower range of $\beta$s for the fixed-KL baselines for fairness.

The primary (sparse) reward for the RoboSumo agent was administered at the end of an episode, with 2000 for a win, -2000 for a loss and -2000 for a draw. To encourage faster learning, as in \cite{al-shedivat2018continuous} and \cite{bansal2018emergent}, we also trained all agents using a dense reward curriculum in the initial stages of training. We refer readers to \cite{al-shedivat2018continuous} for the details of the curriculum, which include auxiliary rewards for agents staying close to the centre of the ring and for being in contact with their opponent. Specifically, for the the first 15\% of training episodes, the agent was given a linear interpolation of the dense and sparse rewards $\alpha r_{dense} + (1-\alpha) r_{sparse}$ with $\alpha$ being decayed linearly from 1 to 0 over the course of the first 15\% of episodes until only the sparse reward was administered. Only the experiences from one of the players in each environment was used to update the agent.

\begin{table*}[h]
\caption{Hyperparameters}
\label{sample-table}
\begin{center}
\begin{small}
\begin{sc}
\begin{tabular}{lcccr}
\toprule
Parameter & Multi-task & Single task & Self-play\\
\midrule
\# Task switches & 19 & 0 & 0 \\
\# Timesteps/Task & 1m & 50m (Humanoid) / 20m (others) & 600m \\
Discount $\gamma$ & 0.99 & 0.99 & 0.995\\
GAE parameter ($\lambda$) & 0.95 & 0.95 & 0.95\\
Horizon & 2048 & 2048 & 8192\\
Adam stepsize (kth policy) & $\omega^{1-k} \times 3 \times 10^{-4}$ & $\omega^{1-k} \times 3 \times 10^{-4}$ or $\omega^{1-k} \times 3 \times 10^{-5}$ & $\omega^{1-k} \times 10^{-4}$\\
VF coefficient & 0.5 & 0.5 & 0.5 \\
\# Epochs per update & 10 & 10 & 6\\
 \# Minibatches & 64 & 64 & 32\\
Neuron type & ReLU & ReLU & ReLU\\
Width hidden layer 1 & 64 & 64 & 64\\
Width hidden layer 2 & 64 & 64 & 64\\
Adam $\beta_1$ & 0.9 & 0.9 & 0.9\\
Adam $\beta_2$ & 0.999 & 0.999 & 0.999\\
\# Hidden policies & 7 & 7 & 7\\
$\omega_{1,2}$ & 1 & 1 & 0.25\\
$\omega$ & 4 & 4 & 4\\
$\beta$ (pol.cons.) & 0.5 & 0.5 & 0.1\\
Adaptive KL $d_{targ}$ & 0.01 & 0.01 & 0.01 \\
\# Environments & 1 & 1 & 16 \\
\bottomrule
\end{tabular}
\end{sc}
\end{small}
\end{center}
\label{table:params}
\end{table*}

\section{Directionality of KL constraint}\label{app:kl}

\begin{figure*}[h]
    \begin{subfigure}[b]{0.5\textwidth}
    \centering
    \includegraphics[width=1.0\textwidth,height=4.3cm]{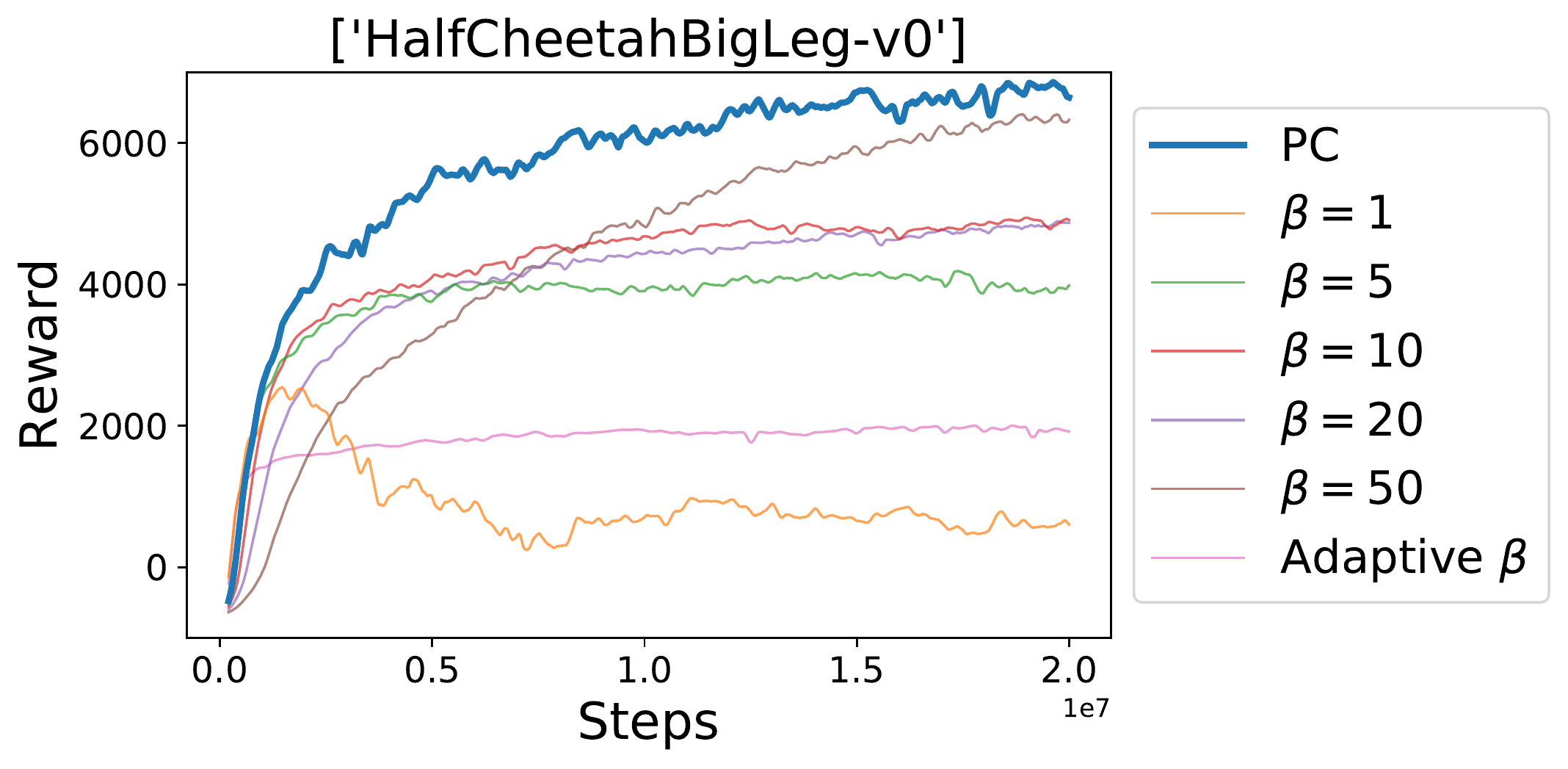}
    \end{subfigure}
    \begin{subfigure}[b]{0.5\textwidth}
    \centering
    \includegraphics[width=1.0\textwidth,height=4.3cm]{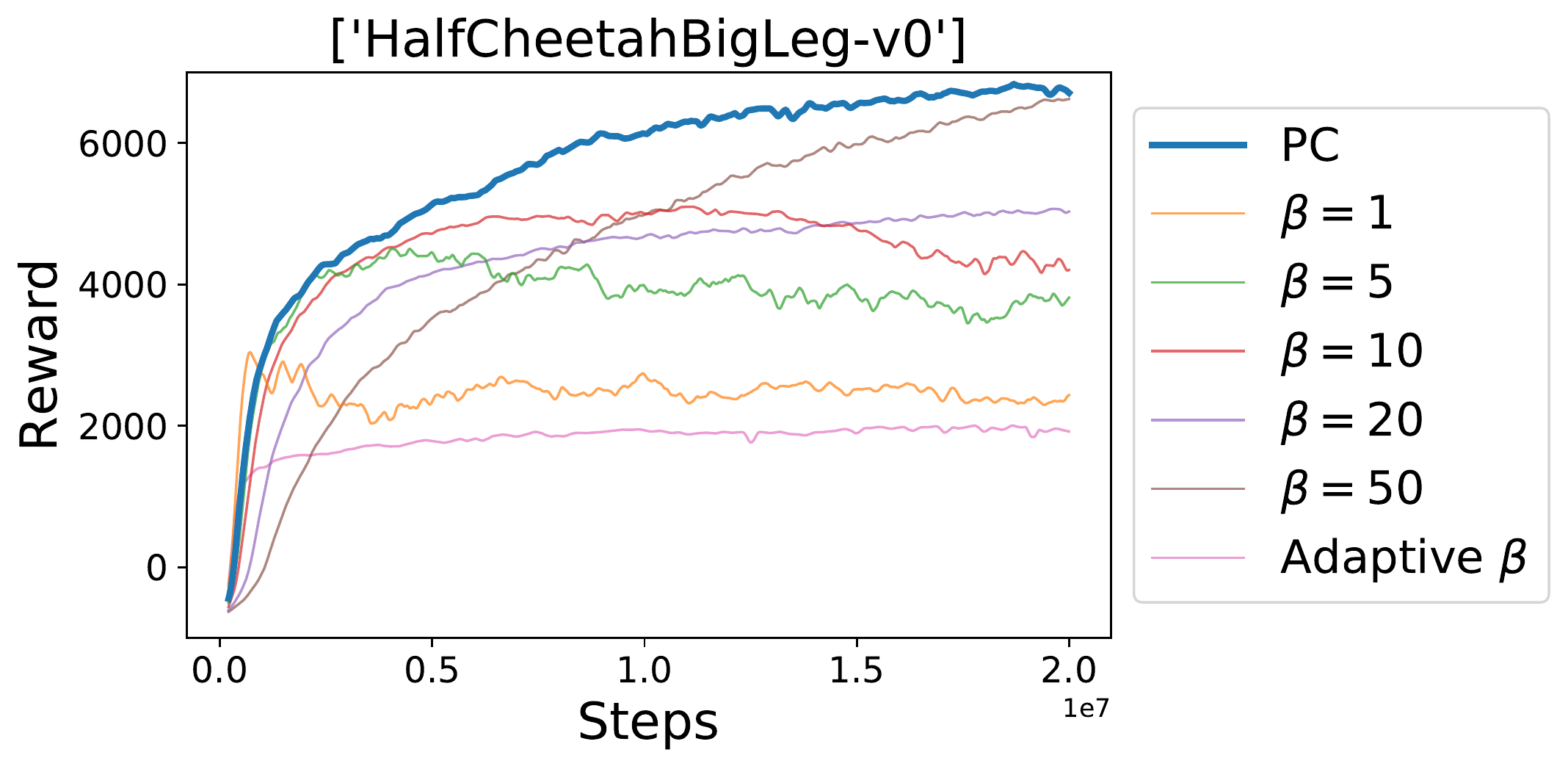}
    \end{subfigure}
    \begin{subfigure}[b]{0.5\textwidth}
    \centering
    \includegraphics[width=1.0\textwidth,height=4.3cm]{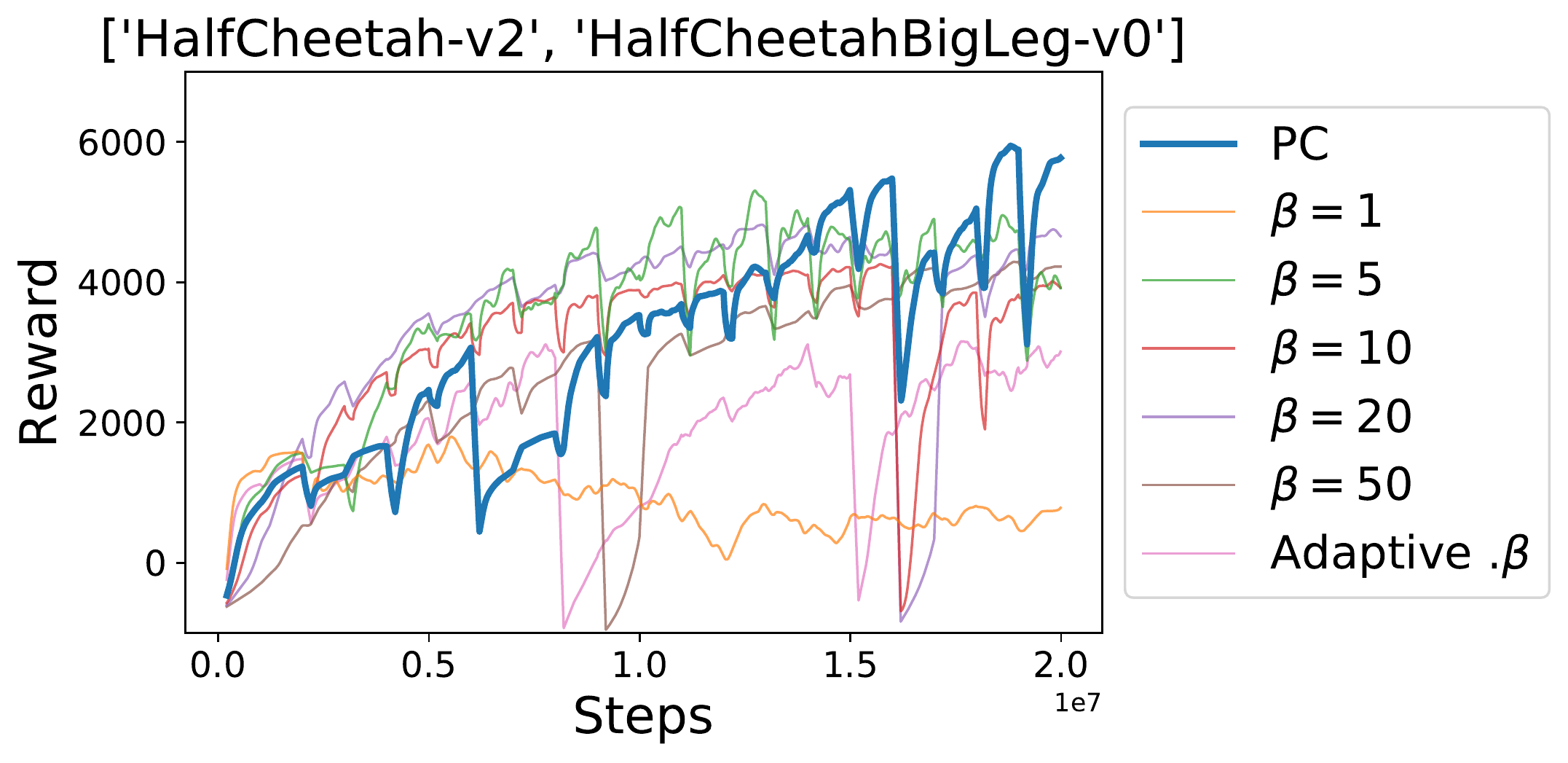}
    \end{subfigure}
    \begin{subfigure}[b]{0.5\textwidth}
    \centering
    \includegraphics[width=1.0\textwidth,height=4.3cm]{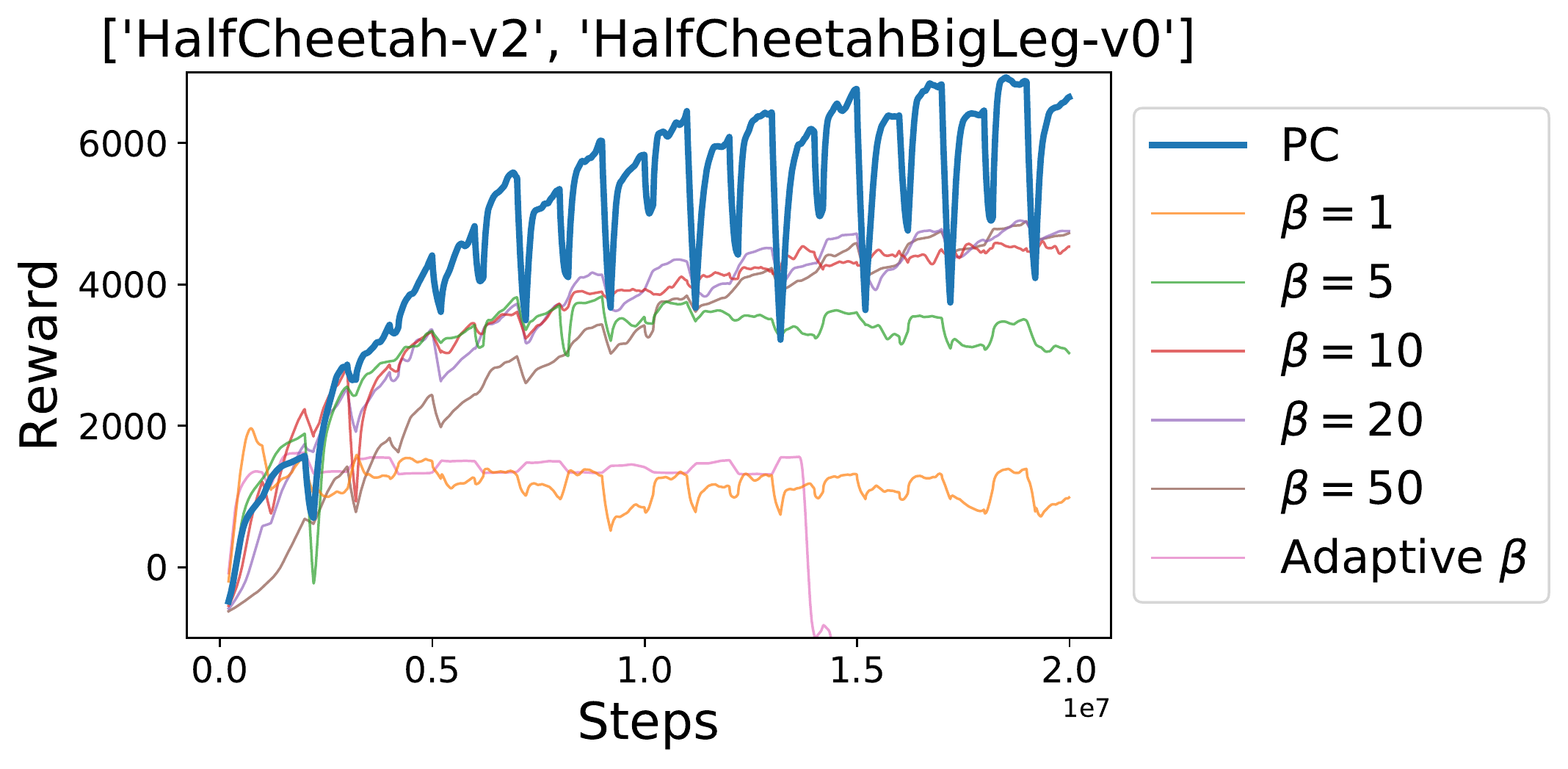}
    \end{subfigure}
    \begin{subfigure}[b]{0.5\textwidth}
    \centering
    \includegraphics[width=1.0\textwidth,height=4.3cm]{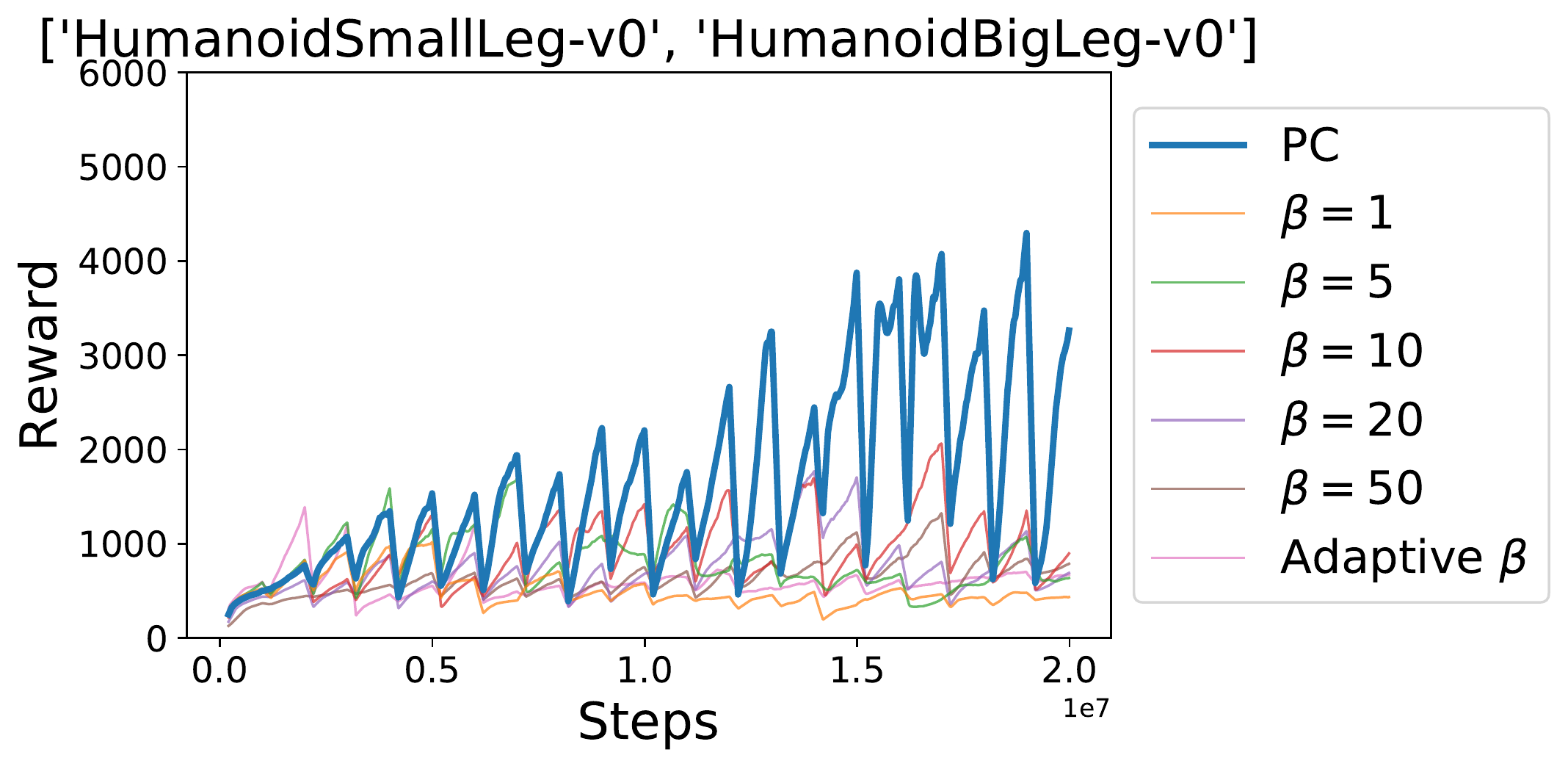}
    \caption{$D_{\mathrm{KL}}\left(\pi_{k_{old}} || \pi_k\right)$}
    \end{subfigure}
    \begin{subfigure}[b]{0.5\textwidth}
    \centering
    \includegraphics[width=1.0\textwidth,height=4.3cm]{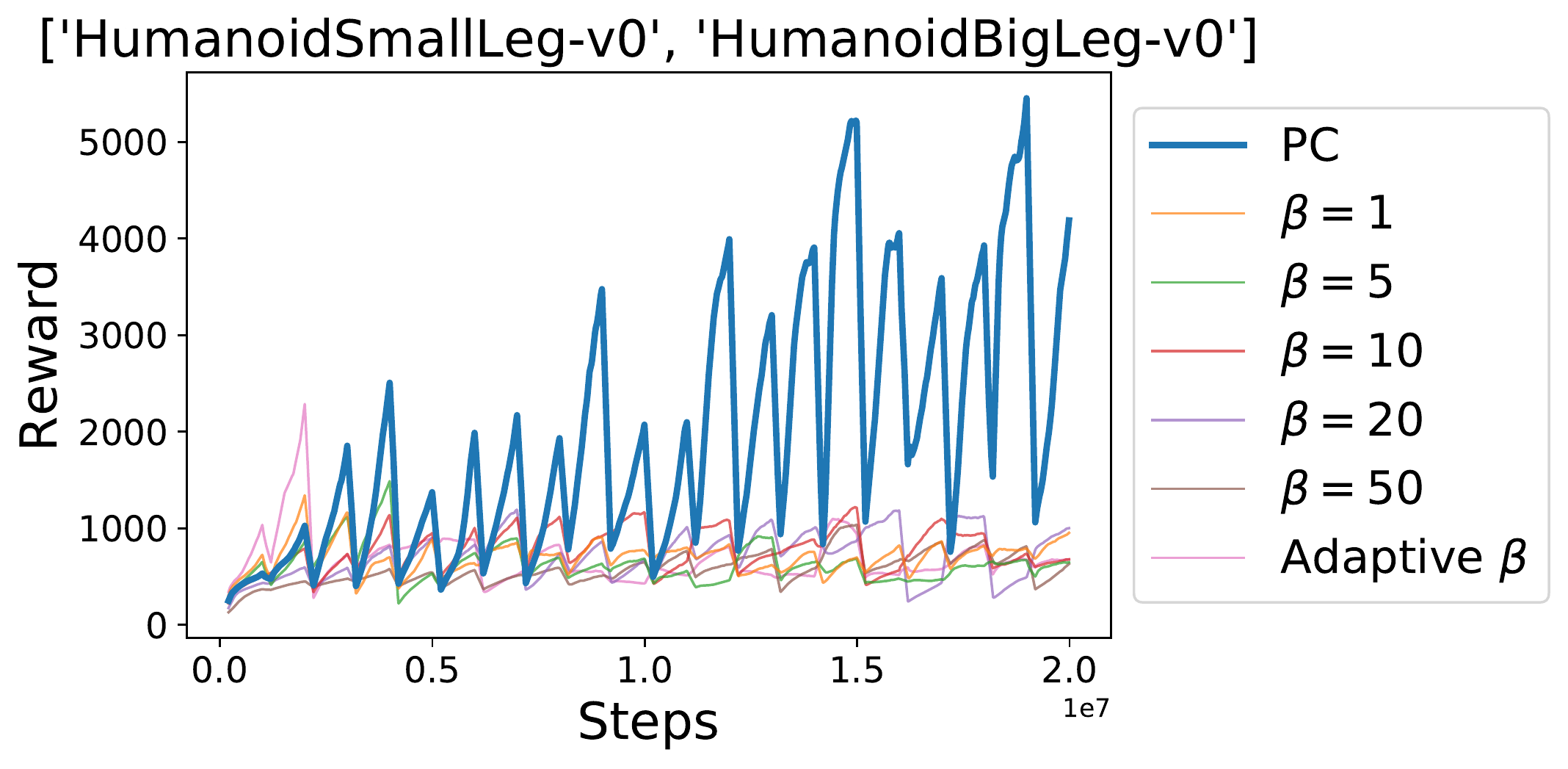}
    \caption{$D_{\mathrm{KL}}\left(\pi_k || \pi_{k_{old}}\right)$}
    \end{subfigure}
    \caption{Reward over time using the (a) $D_{\mathrm{KL}}\left(\pi_{k_{old}} || \pi_k\right)$ and (b) $D_{\mathrm{KL}}\left(\pi_k || \pi_{k_{old}}\right)$ constraints.}
    \label{fig:kl}
\end{figure*}

In our initial experiments we found that using a  $D_{\mathrm{KL}}\left(\pi_k || \pi_{k_{old}}\right)$ constraint for each policy in the PC model, rather than the $D_{\mathrm{KL}}\left(\pi_{k_{old}} || \pi_k\right)$ constraint used in the KL versions of PPO \cite{schulman2017proximal}, resulted in better continual learning and so in the main results section we compared the PC model with KL baselines that also used the $D_{\mathrm{KL}}\left(\pi_k || \pi_{k_{old}}\right)$ constraint.  Here we show in a few experiments that we get the same qualitative improvements from the PC agent if we use the original KL constraint from PPO for both the PC model and the baselines (Figure \ref{fig:kl}). As can be seen particularly in the HalfCheetah and Humanoid alternating task settings, the $D_{\mathrm{KL}}\left(\pi_k || \pi_{k_{old}}\right)$ version performs better.

The effect of the directionality of this KL constraint, as well as the directionality of the KL constraints between adjacent policies (of which there are four possible combinations) warrants further investigation and is an important avenue for future work.

\section{Task switching schedule effects}
\begin{figure*}[h]
    \centering
    \begin{subfigure}{0.49 \textwidth}
    \centering
    \includegraphics[width=\textwidth]{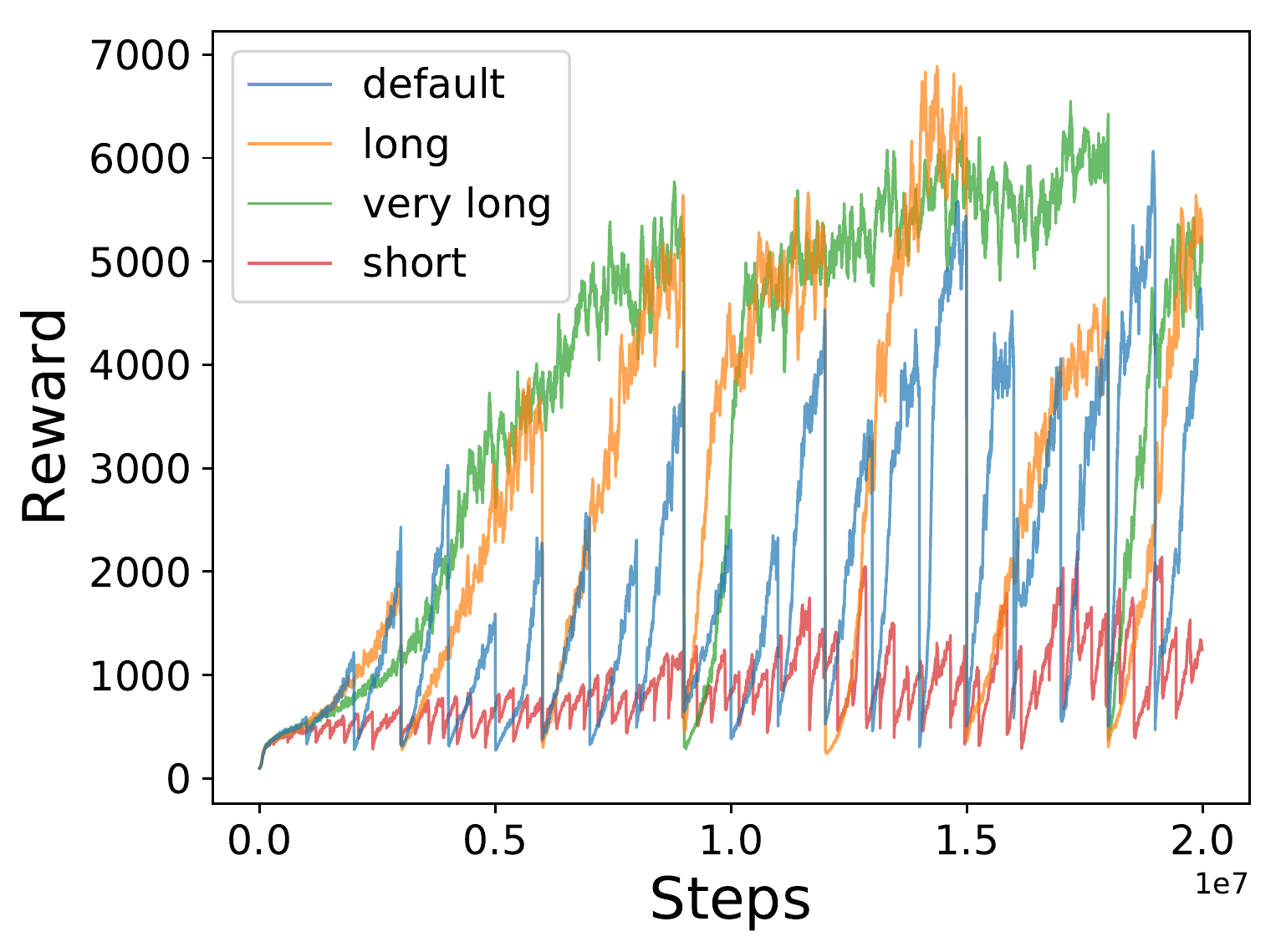}
    \caption{PC model}
    \end{subfigure}
    \begin{subfigure}{0.49 \textwidth}
    \centering
    \includegraphics[width=\textwidth]{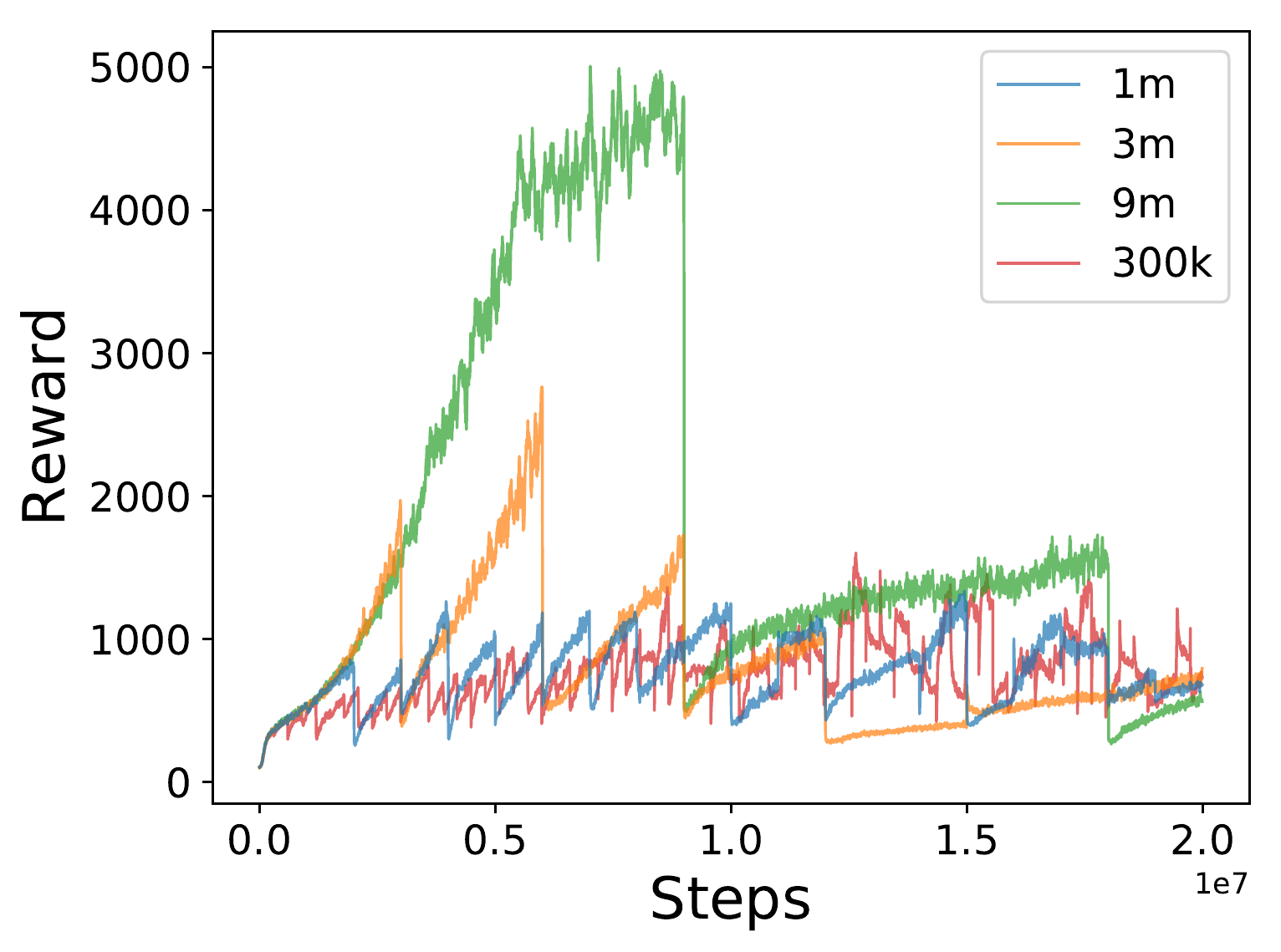}
    \caption{Fixed-KL $\beta=10$}
    \end{subfigure}
    \caption{Reward over time for (a) PC model and (b) fixed-KL baseline with $\beta=10$ for different task-switching schedules between the HumanoidSmallLeg-v0 and HumanoidBigLeg-v0 tasks.}
    \label{fig:task_time}
    \vspace{-0.5cm}
\end{figure*}

Figure \ref{fig:task_time} shows the effects of changing the frequency of task switching in the alternating task setting for both the PC model and one of the baselines (fixed-KL with $\beta=10$). An interesting point to note is that in the baseline runs with slower task-switching schedules, the performances on both tasks decrease over time, with the agent unable to reach previously attained highs. In other words, the agent not only catastrophically forgets, but learning one task puts the network in a state that it struggles to (re)learn the other task at all.


\end{document}